\pdfoutput=1
\documentclass[runningheads]{llncs}
\usepackage{graphicx}
\usepackage{tikz}
\usepackage{comment}
\usepackage{amsmath,amssymb}
\usepackage{color}
\usepackage{epsfig}
\usepackage{enumitem}
\usepackage{booktabs}
\usepackage{multirow}
\usepackage{floatrow}
\usepackage[ruled]{algorithm2e}
\usepackage[pagebackref=true,breaklinks=true,colorlinks,bookmarks=false]{hyperref}
\renewcommand\thefootnote{}
\usepackage[accsupp]{axessibility}
\newfloatcommand{capbtabbox}{table}[][\FBwidth]

\begin{document}
\pagestyle{headings}
\mainmatter
\title{Facial Geometric Detail Recovery via\\ Implicit Representation}

\titlerunning{Facial Geometric Detail Recovery via Implicit Representation}

\author{Xingyu Ren\inst{1} \and
Alexandros Lattas\inst{2,3} \and
Baris Gecer\inst{3} \and
Jiankang Deng\inst{3} \and \\
Chao Ma\inst{1} \and
Xiaokang Yang\inst{1} \and
Stefanos Zafeiriou\inst{2,3}}

\authorrunning{X. Ren, A. Lattas, B. Gecer, J. Deng, C. Ma, X. Yang, S. Zafeiriou}
\institute{MoE Key Lab of Artificial Intelligence, AI Institute, Shanghai Jiao Tong University
\email{\{rxy\_sjtu, chaoma, xkyang\}@sjtu.edu.cn} \and
Imperial College London \email{\{a.lattas,b.gecer,j.deng16,s.zafeiriou\}@imperial.ac.uk}
\and Huawei CBG}

\maketitle
\begin{abstract}
Learning a dense 3D model with fine-scale details from a single facial image is highly challenging and ill-posed.
To address this problem, many approaches fit smooth geometries through facial prior while learning details as additional displacement maps or personalized basis.
However, these techniques typically require vast datasets of paired multi-view data or 3D scans, whereas such datasets are scarce and expensive.
To alleviate heavy data dependency, we present a robust texture-guided geometric detail recovery approach using only a single in-the-wild facial image.
More specifically, our method combines high-quality texture completion with the powerful expressiveness of implicit surfaces.
Initially, we inpaint occluded facial parts, generate complete textures, and build an accurate multi-view dataset of the target subject.
In order to estimate the detailed geometry, we define an implicit signed distance function and employ a physically-based implicit renderer to reconstruct fine geometric details from the generated multi-view images.
Our method not only recovers accurate facial details but also decomposes the normals, albedos and shading components in a self-supervised way. Finally, we register the implicit shape details to a 3D Morphable Model template, which can be used in traditional modeling and rendering pipelines.
Extensive experiments demonstrate that the proposed approach can reconstruct impressive facial details from a single image, especially when compared with state-of-the-art methods trained on large datasets.
The source code is available at \url{https://github.com/deepinsight/insightface/tree/master/reconstruction/PBIDR}.
\footnote{This work is done when Xingyu Ren is an intern at Huawei.} 
\setcounter{footnote}{0}
\renewcommand\thefootnote{\arabic{footnote}}
\keywords{Geometric Detail Recovery, Texture Completion, Implicit Surface}
\end{abstract}

\section{Introduction}
\begin{figure}[t]
    \centering
    \includegraphics[width=\linewidth]{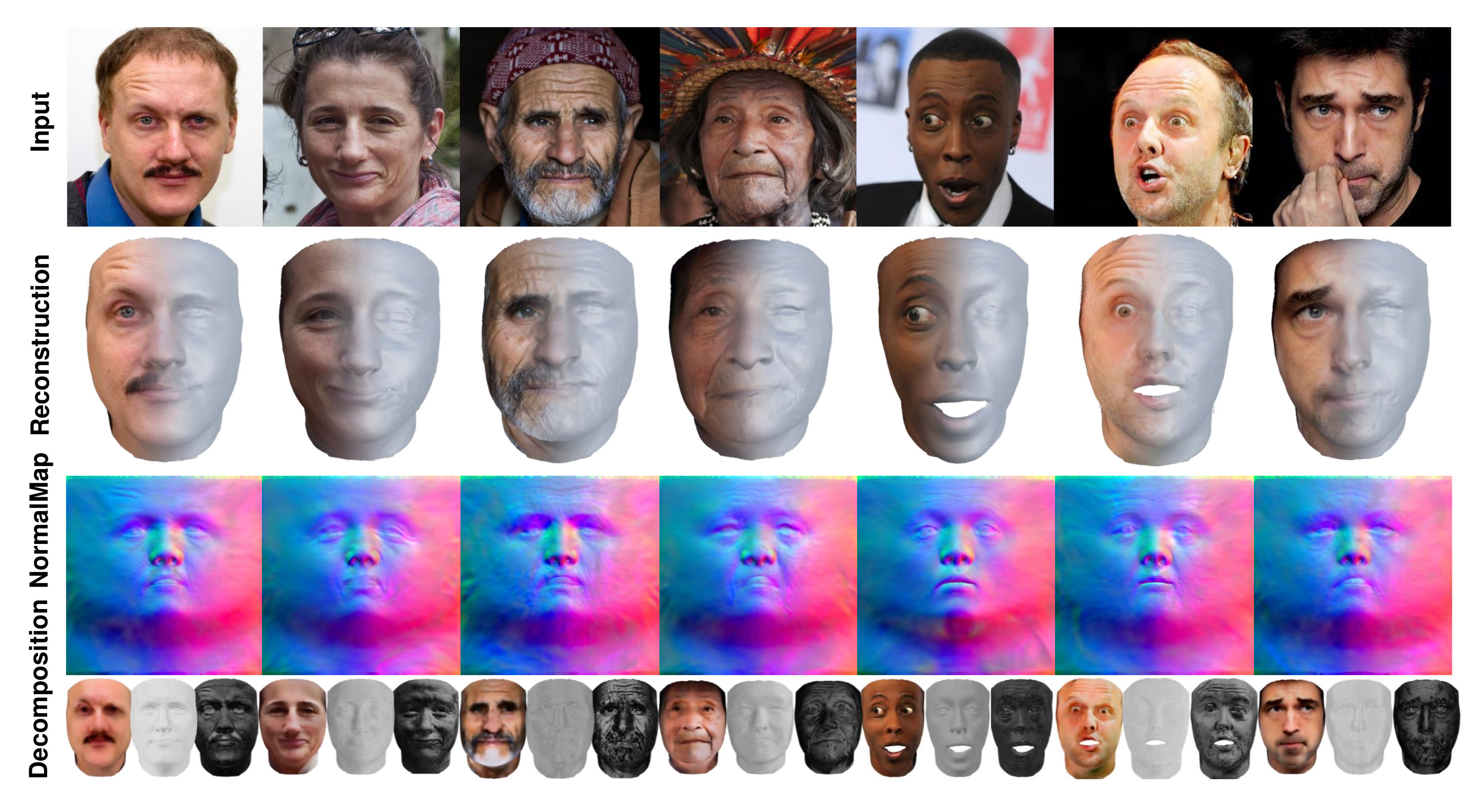}
    \caption{\small We introduce a single-facial-image geometric detail recovery algorithm. Our method generates complete high-fidelity texture maps from occluded facial images, and employs implicit renderer and shape functions, to derive fine geometric details from decoupled detail normals. 
    As a bonus, it disentangles the facial texture into approximate diffuse albedo, diffuse and specular shading in a self-supervision manner. Input images are from FFHQ~\cite{karras2019style} and CelebAMask-HQ~\cite{CelebAMask-HQ}.}
    \label{fig:Figure1}
\end{figure}

Estimating realistic 3D human face shapes from 2D images is a challenging and ill-posed problem in computer vision and graphics, which requires strong prior knowledge to reduce appearance and depth ambiguities. Since the seminal work of 3D Morphable Model (3DMM)~\cite{vetter19983dmm},
monocular 3D face reconstruction methods have achieved impressive results in many applications, e.g., face recognition \cite{blanz2002face}, manipulation \cite{thies2016face2face}, and speech-driven facial animation \cite{thies2020neural,tzirakis2020synthesising}. However, such limited low-dimensional representations \cite{chang2018expnet,genova2018unsupervised,kim2018inversefacenet,sanyal2019learning,deng2019accurate,ploumpis2020towards}
omit shape details, such as wrinkles, and result in smooth reconstructions.

To tackle this problem, some approaches \cite{richardson2017learning,guo2018cnn,tran2018extreme,chen2019photo,bai2020deep,gecer2020synthesizing,feng2021learning} propose to improve 3DMM representation capacity by adding non-linearity, while others \cite{sengupta2018sfsnet,abrevaya2020cross,gecer2020synthesizing} even directly learn to model faces without 3DMM bases. However, all the above methods require costly training data, such as paired multi-view faces or high fidelity 3D scans. 
In this paper, we aim at recovering accurate geometric details from a single image, without relying on large datasets of 3D facial scans.

For single image based face reconstruction, a critical issue is how to effectively represent geometric details.
Current state-of-the-art approaches \cite{yang2020facescape,feng2021learning,lattas2021avatarme++} utilize neural networks to generate displacement or normal maps to model facial details.
However, such a representation is only suitable for large datasets and tends to produce over-smooth results.
Tewari et al. \cite{tewari2018self} explicitly update vertices on coarse meshes but hardly obtain fine-scale wrinkles as their pre-defined vertices are sparse.
Some other approaches explore facial appearance as a photometric loss to calculate normals and depth, such as the well-known ``shape-from-shading" method \cite{jiang20183d,li2018feature,riviere2020single,li2018feature}.
Unfortunately, using a single image suffers from missing appearance features, such as occlusion from foreign objects or self-occlusion under a large head pose.
Even more, the ambiguity from environmental factors, such as shading, makes the facial shape reconstruction an ill-posed problem.
Although some methods (i.e.,~\cite{smith2020AlbedoMM,deng2019accurate}) involve texture prior as a parametric texture space, such low-dimensional representations hardly reconstruct fine-scale textures with mesoscopic facial details.
Our insight is that
implicit surfaces contain flexible resolutions and meaningful geometric details, which can be extracted by aligned meshes in a self-supervised way.

In this paper, we propose a novel facial geometric detail recovery approach, which includes a robust texture completion using a style-based facial generator and a geometry enhancement using an implicit differentiable rendering optimization. 
For the texture completion component, we first segment the occluded facial areas and then utilize a pre-trained style-based generator to restore the object-occluded parts.
Specifically, an occlusion-free 2D face image is first generated from StyleGANv2 \cite{Karras2019stylegan2} by taking the masked image as the input. Then, the masked image is in-painted by the generated occlusion-free 2D face. In this way, the object-occluded regions are restored while other regions are exactly from the original image.
After removing the external occlusions, we follow a similar procedure to recover the self-occluded facial areas and generate complete, clean, and high-fidelity  UV texture maps. 

For the geometry enhancement component, we utilize the completed facial texture and a coarse shape fitting to render multi-view facial images.
Motivated by the recent advancements in geometry reconstruction with implicit functions \cite{mescheder2019occupancy,peng2020convoccupancy,chen2019imnet}, we improve the coarse geometry using the high-fidelity multi-view facial renderings.
In that manner, we propose a physically-based implicit differentiable rendering framework, which approximates a bidirectional reflectance distribution (BRDF) function for facial images.
Our framework separates diffuse and specular normals under coarse fitting constraints, obtaining rich geometric details using iterative optimization from the rendered images.
Furthermore, we propose a registration loss to predict the signed distance between the implicit zero-order set and the coarse mesh. In this way, geometric details are explicitly transferred back to the coarse 3DMM topology.

To verify the effectiveness of the proposed geometric detail recovery approach,
we conduct exhaustive evaluations on NoW \cite{sanyal2019learning}, MICC Florence \cite{bagdanov2011florence}, 3DFAW \cite{pillai20192nd} and our collected 3D scans.
The results show that our method consistently achieves competitive performance compared to state-of-the-art methods in shape reconstruction.
Moreover, our approach outperforms the current geometric detail recovery methods, achieving the lowest normal cosine distance errors on our collected dataset.
To summarize, our major contributions are as follows:
\begin{itemize}[noitemsep,nolistsep]
\item We present a novel method for extracting high-fidelity texture and detailed facial geometry from a single image, without requiring vast facial images or 3D scans.
\item We design an occlusion-robust facial texture completion method, using only a pre-trained StyleGANv2 generator.
\item We propose a detailed shape optimization method, using a novel physically-based implicit differentiable rendering function, which decouples specular normals for precise details recovery.
\item Our qualitative and quantitative experiments have shown that the proposed approach obtains competitive results on shape reconstruction and achieves state-of-the-art detail recovery performance on our collected dataset.
\end{itemize}
\section{Related Work}

\noindent{\bf Texture Completion.}
3D face modeling and synthesis is a widely researched subject \cite{gecer2019ganfit,slossberg2018high,saito2017photorealistic}.
Recent works aim to complete high-fidelity 3D texture maps from 2D images \cite{deng2018uv,gecer2019ganfit,kim2021learning,gecer2021ostec}.
Deng et al. \cite{deng2018uv} utilize face symmetry to train an image-to-image translation network to generate full texture, heavily requiring selected training data.
Gecer et al. \cite{gecer2021ostec} treat texture completion as an inpainting problem and utilize StyleGANv2 \cite{Karras2019stylegan2} to synthesize self-occlusion parts.
Unfortunately, the above methods cannot restore correct texture maps when external occlusions exist, such as glasses, hands, etc.

\noindent{\bf Detailed Geometry Recovery.}
Methods of restoring geometric details can be divided into optimization-based \cite{jiang20183d,li2018feature,garrido2016reconstruction}, regression-based \cite{chen2019photo,cao2015real,richardson2017learning,guo2018cnn,tran2018extreme,bai2020deep,feng2021learning}, and model-free \cite{tran2019towards,tewari2018self,tewari2019fml,lattas2020avatarme,lee2020uncertainty} methods. The optimization-based methods first fit the 3DMM model to get coarse geometry and then use the well-known ``shape from shading'' \cite{ramachandran1988perception} approach to generating details. However, they are sensitive to occlusions and can only be performed in the visible areas.
Similarly, regression-based methods also try to fit a rough shape and refine it in various ways. Cao et al. \cite{cao2015real} and Chen et al. \cite{chen2019photo} employ local statistics from high-resolution scans to reconstruct fine detail meshes. Richardson et al. \cite{richardson2017learning} and Guo et al. \cite{guo2018cnn} directly derive per-pixel displacement maps. Tran et al. \cite{tran2018extreme} regress a bump map, and gain robustness to occlusions by applying a face segmentation method. Bai et al. \cite{bai2020deep} propose a 3D face reconstruction from multi-view images with different expressions by a non-rigid multi-view stereo optimization framework while simultaneously requiring high-quality 3d scans for training. Lattas et al. \cite{lattas2021avatarme++} use image-to-image translation networks to learn diffuse and specular reflectance using light-stage data. Feng et al. \cite{feng2021learning} directly learn an animatable displacement map and obtain a robust shape when occlusions exist. While the above require large amounts of training data, we fit a coarse mesh and then directly learn geometric details from a single image by implicit functions.

Moreover, model-free methods directly reconstruct dense mesh or infer surface normals that add to coarse reconstructions. Tran et al. \cite{tran2019towards} and Tewari et al. \cite{tewari2018self,tewari2019fml,tewari2018high} propose a learnable face model to reconstruct 3D faces from images.
Although they have achieved realistic rendering results and lots of details in texture, limited geometric details are captured.

\noindent{\bf Implicit Surface Reconstruction.}
Recently, implicit representations have achieved impressive results when modeling surfaces as a level-set of a coordinate-based continuous function, e.g., a signed distance function \cite{park2019deepsdf} or an occupancy function \cite{mescheder2019occupancy,peng2020convoccupancy,chen2019imnet}. We focus on implicit surface reconstruction from point clouds \cite{atzmon2020sal,atzmon2020sald,NeuralPull,chibane2020NUDF} and 2D images \cite{liu2020dist,jiang2020sdfdiff,niemeyer2020dvr,yariv2020idr}. Common approaches involve additional geometric \cite{gropp2020igr} and gradient \cite{atzmon2020sald,chibane2020NUDF,NeuralPull} constraints to enforce implicit function compact when learning from raw point clouds without ground truth. Two recent papers Neural-Pull \cite{NeuralPull} and NUDF \cite{chibane2020NUDF} predict signed and unsigned distances, respectively, and calculate the implicit surface by pulling the sampled query to the surface. For an input image, approaches like \cite{liu2020dist,jiang2020sdfdiff,niemeyer2020dvr,yariv2020idr} employ a differentiable renderer and optimize an implicit function to learn the surface. 
Even though the implicit differentiable renderer (IDR) \cite{yariv2020idr} exhibits impressive results on inanimate objects, this framework is inadequate to capture the complexity of human skin geometry and shading.
The most relevant to us is SIDER \cite{chatziagapi2021sider}, which first encodes a reconstructed flame-based mesh to the implicit surface and then utilizes a single image to recover details in implicit spaces. However, such a single view setting causes unrealistic facial details and cannot be driven by other expressions. 

\section{Methods}

The principal intuition behind this study is to exploit the advances in face generators and implicit rendering, in order to reconstruct detailed facial shape from a single image. Specifically, we first employ an off-the-shelf 3DMM face reconstruction algorithm~\cite{gecer2019ganfit} and propose an occlusion-robust texture completion method built on~\cite{gecer2021ostec} to obtain an initial coarse geometry, de-occluded high-resolution texture map and re-rendered images at various views (Sec.~\ref{texture3.1}).

We then present a self-supervised optimization pipeline based on an implicit signed distance function and physically-based implicit differentiable renderer to decompose diffuse normal and specular normal for detailed geometry recovery (Sec.~\ref{Reconstruction3.2}).
As a bonus, we can disentangle diffuse albedo, diffuse shading and specular shading from the acquired texture map in a self-supervised way (Sec.~\ref{Reconstruction3.2}). 
Finally, the estimated shape details are embossed to the coarse mesh by registration and local displacement (Sec.~\ref{Transfer3.2.4}).

\begin{figure}[t]
\centering
\includegraphics[width=\linewidth]{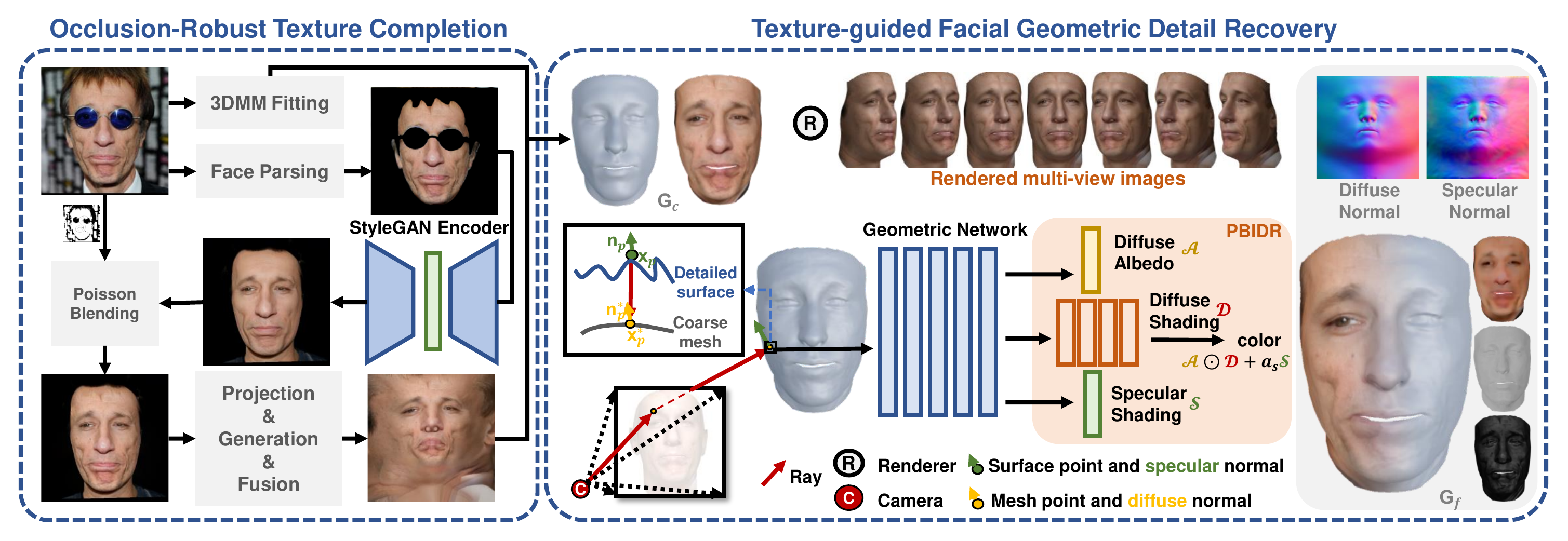}
 \vspace{-0.2cm}
\caption{\small Overview of our method. Left: we perform a robust texture completion, utilizing StyleGANv2 \cite{Karras2019stylegan2} to repair both externally- and self-occluded regions. Right: Based on a coarse 3DMM fitting, we render a multi-view dataset and use an implicit signed distance function and a novel physically-based implicit differentiable renderer (PBIDR) to improve the coarse shape. Based on the self-supervised decomposition of diffuse normal and specular normal, we obtain a detailed facial geometry as well as approximate diffuse albedo, diffuse shading and specular shading.}
 \vspace{-0.4cm}
\label{fig:figure2_pipeline}
\end{figure}

\subsection{Occlusion-Robust Texture Completion}\label{texture3.1}
Given an input 2D image $\mathbf{I}_{i}$, we aim to restore a high-quality texture map $\mathbf{T}_{o}$ by recovering externally occluded and self-occluded parts of the facial appearance.
To this end, we
(1) estimate a segmentation mask of the object-occluded regions and inpaint them in the original image space by projecting the masked face image into the StyleGANv2 latent space,
(2) complete the invisible texture by rotating 3D facial geometry and inpainting the self-occluded areas in the UV space \cite{gecer2021ostec}, and
(3) re-render the 3D face with the coarse geometry and the completed UV texture at various views to be used in Sec.~\ref{Reconstruction3.2}.
\vspace{-3mm}
\subsubsection{Removing External Occlusions}\label{sec:remove_occlusion}
We first remove external occlusions, such as accessories, hands, hair etc., by inpainting those segmented areas in the following manner. We employ a high-resolution 2D face generator as a strong face prior, StyleGANv2 \cite{Karras2019stylegan2}, which 
can generate desired high-quality faces $\mathbf{I}_{r}$ by sampling the latent space.
Given the input image $\mathbf{I}_{i}$ and the segmentation mask $\mathbf{M}_{i}$ acquired by an off-the-shelf facial segmentation model~\cite{lin2021roi}, we propose a self-fusion approach, which simultaneously (1) minimizes the discrepancy between the reconstructed image $\mathbf{I}_{r}$ and the input image $\mathbf{I}_{i}$,  and (2) maximizes the visual plausibility of the final fused image $\mathbf{I}_{f}$($=\mathbf{M}_{i}\mathbf{I}_{i} + (1-\mathbf{M}_{i})\mathbf{I}_{r}$), as shown in Fig.~\ref{fig:figure2_pipeline}.
Note that $\mathbf{I}_{r}$ and $\mathbf{I}_{f}$ have different optimization objectives. $\mathbf{I}_{r}$ aims at exactly reconstructing the visible parts of the original image $\mathbf{I}_{i}$ and $\mathbf{I}_{f}$ targets on generating coherent content for occluded parts with the plausible overall style. If the semantics of $\mathbf{I}_{f}$ were over-constrained, our pipeline would have limited capacity, generating unexpected occlusions similar to the input image.
After the whole optimization is completed, we utilize the Poisson Blending to achieve a high-fidelity post-fusion result $\mathbf{I}_{f}$.

For the loss design, we employ
photometric loss $\mathcal{L}_{photo}$, 
landmark loss $\mathcal{L}_{pts}$, and
identity loss $\mathcal{L}_{id}$, 
similarly to \cite{gecer2021ostec}.
We also add 
perceptual loss $\mathcal{L}_{per}$
and discriminator loss $\mathcal{L}_{dis}$ 
for restored image $\mathbf{I}_{r}$, 
as well as a MS-SSIM loss $\mathcal{L}_{mssim}$ for $\mathbf{I}_{f}$ only. Therefore, the overall objective function is as following: 
\begin{equation}\label{eq1}
     \mathcal{L}_{total}=\lambda_{1;r}\mathcal{L}_{photo}+\lambda_{2;r}\mathcal{L}_{pts} + \lambda_{3;r}\mathcal{L}_{id} + \lambda_{4;r}\mathcal{L}_{per} + \lambda_{5;r}\mathcal{L}_{dis} +\lambda_{6;f}\mathcal{L}_{mssim},
\end{equation}
where $\lambda_{i;r}$ is used by $\mathbf{I}_{r}$ and $\lambda_{i;f}$ is for $\mathbf{I}_{f}$.
\vspace{-3mm}
\subsubsection{Texture Completion}
After removing external occlusions and post-fusion, we obtain a cleaned face $\mathbf{I}_{f}$. 
Following OSTeC \cite{gecer2021ostec}, we apply an off-the-shelf 3D reconstruction method~\cite{gecer2019ganfit} to estimate a coarse 3DMM shape fitting ($\mathbf{G}_{c}$), and then acquire visible parts of the textures from $\mathbf{I}_{f}$, re-render at different poses and inpaint the self-occluded parts at different views using a similar approach as explained in Sec.~\ref{sec:remove_occlusion}. By blending generated textures one by one, we achieve a high-fidelity texture $\mathbf{T}_{o}$. Unlike OSTeC \cite{gecer2021ostec}, our texture completion approach is not affected by occlusions, producing clean completed textured maps.
\vspace{-3mm}
\subsubsection{Re-rendering of Multi-view Faces}
Finally, we render the mesh $\mathbf{G}_{c}$ with the texture map $\mathbf{T}_{o}$ at various preset camera poses, in order to gather a set of multi-view images of the subject
$\mathbf{I} ^{\ast } = \{\mathbf{I} ^{v}, v =1,2,\dots ,17\}$ to be used in the following section.

\subsection{Texture-guided Facial Geometric Detail Recovery}\label{Reconstruction3.2}
Given a set of high-fidelity and consistent renderings of the subject $\mathbf{I} ^{\ast }$ and their corresponding coarse geometry $\mathbf{G}_{c}$, 
we define a signed distance function (SDF) (Sec.~\ref{Shape3.2.1})
and a Physically-Based Implicit Differentiable Renderer (PBIDR) (Sec.~\ref{Rendering3.2.2}), motivated by \cite{yariv2020idr}.
We then optimize the SDF using PBIDR (Sec.~\ref{Optimization3.2.3}) in order to obtain precise specular normals and recover a detailed implicit facial surface, which is finally transferred back to the 3DMM topology (Sec.~\ref{Transfer3.2.4}).
\vspace{-4mm}
\subsubsection{Shape Representation as an SDF}\label{Shape3.2.1}
Here, we aim to optimize an SDF that describes the detailed shape of the reconstructed subject, based on the images of that subject $\mathbf{I} ^{\ast }$. 
To this end, we define the SDF and the ray-SDF intersection as follows.
We formulate an SDF such that $\mathcal{F}_{\theta }: \mathbf{x}  \in  \mathbb{R}^{3} \rightarrow s  \in  \mathbb{R}^{1}$ to approximate the implicit surface $\mathcal{G}= \left \{ \mathbf{x} \mid  \mathcal{F} \left ( \mathbf{x} ;\mathrm {\theta }\right )=0  \right \} $. 
Also, $\mathcal{F}_{\theta}$ generates a global geometry feature vector $\mathbf{z_{d}}$,
that captures global effects.
Similarly to \cite{yariv2020idr}, $\mathcal{F}_{\theta }$ is learned with an MLP.
For each pixel $p$ in auxiliary images $\mathbf{I} ^{\ast }$, we utilize ray marching to obtain the intersection $\mathbf{x}_{p}$ between rays and implicit surface via the sphere tracing \cite{jiang2020sdfdiff} algorithm. 
The intersection is then learned by adding a layer before and after the neural network $\mathcal{F}$,
that encodes the SDF.
We additionally obtain another intersection $\mathbf{x}^{\ast}_{p}$ between rays and coarse mesh $\mathbf{G}_{c}$. 
\vspace{-4mm}
\subsubsection{Physically-Based Implicit Differentiable Renderer}\label{Rendering3.2.2}
Prior art \cite{yariv2020idr} uses a neural network to approximate the rendering equation $\mathcal{R}_{\varphi}$, with impressive results on objects with simple reflectance.
However, we find this framework inadequate to capture the complexity of human skin, whose reconstruction result suffers from high-frequency noises and artifacts (as shown in Fig.~\ref{fig:Figure6_multi_view_mesh}).
Instead, we introduce the following Physically-Based Implicit Differentiable Renderer (PBIDR).

Traditionally, facial reconstruction methods \cite{deng2019accurate}
use a simplistic Lambertian shading model to render a facial mesh with a single albedo texture.
On the contrary, realistic facial rendering requires additional material properties
and expensive shading models.
The Blinn-Phong shading model \cite{phong1975illumination}
separately models diffuse and specular shading,
and can approximate human skin rendering at a low computational cost.
Moreover, separating the diffuse and specular normals
can also approximate the skin's diffuse subsurface scattering \cite{lattas2021avatarme++},
which is less affected by the high-frequency details present in the specular normals.
On the contrary, specular normals represent the appearance of some mesoscopic surface details, such as fine wrinkles and skin pores, which are often challenging to extract, but their effect is significant in rendering \cite{lattas2021avatarme++}.

We approximate the detailed specular normals as the gradient of our detailed SDF shape $\mathbf{n}_{p} = \nabla_\mathbf{x}\mathcal{F}_{\theta}(\mathbf{x}_{p})$
and the smooth diffuse normals $\mathbf{n}^{\ast}_{p}$ as the normal obtained by barycentric sampling on the original coarse mesh $\mathbf{G}_{c}$, and a prior is $\mathbf{n}_{p}$ and $\mathbf{n}^{\ast}_{p}$ are not far apart in direction.
In this manner, only the SDF normals are responsible for the high-frequency details of the rendered image, guiding the SDF to capture them more accurately.
Therefore, we decompose a subject's appearance at a point $p$ to\\
(1) a diffuse albedo $\mathcal{A}(\mathbf{x}_{p}, \mathbf{z}_{p})\in\mathbb{R}^3$ 
as a function of the surface point $\mathbf{x}_{p}$ and the feature vector $\mathbf{z}_{p}$,
(2) diffuse shading 
$\mathcal{D}(\mathbf{n}^{\ast}_{p})\in\mathbb{R}$
as a function of diffuse normals $\mathbf{n}^{\ast}_{p}$,
(3) specular albedo $a_s \in \mathbb{R}$, which is assumed a constant to decrease ambiguity in the optimization, and (4) a specular shading 
$\mathcal{S}(\mathbf{n}_{p}, \mathbf{v})\in\mathbb{R}$
as a function of specular normals $\mathbf{n}_{p}$ and the view direction $\mathbf{v}$,
since the specular shading is view-dependent.
Except for the constant $a_s$, we use an MLP to encode each of these properties which are trained in an unsupervised way during the optimization.
Finally, the complete rendering function is defined as:
\begin{equation}\label{eq2}
 \mathbf{c}_{p} ( \mathbf{v}) = \mathcal{A}(\mathbf{x}_{p}, \mathbf{z}_{p}) \mathcal{D}(\mathbf{n}^{\ast}_{p}) + a_s \; \mathcal{S}(\mathbf{n}_{p}, \mathbf{v}).
\end{equation}
\vspace{-8mm}
\subsubsection{Shape Details Optimization}\label{Optimization3.2.3}
We define an optimization framework and train the SDF MLP $\mathcal{F}_{\theta }$
using photometric loss, masking loss,
eikonal loss,
registration loss (Eq.~\ref{Eq_L_reg}),
and normal loss (Eq.~\ref{Eq_L_nor}).

Let $\mathcal{P}$ be a mini-batch of pixels from view direction $\mathbf{v}$ and image $\mathbf{I} ^{v}$ in the training process, $\mathbf{M}^{v}$ is a mask of rendered parts of $\mathbf{I} ^{v}$, and $\mathcal{P}^{rgb}$ is a subset of $\mathcal{P}$ where intersections $\mathbf{x}_{p}$ and $\mathbf{x}^{\ast}_{p}$ projects on. We predict a final color value $\mathbf{c}_{p},p\in\mathcal{P}$ as described in the above, and define a photometric loss $\mathcal{L}_{photo}$ between pixel $\mathbf{p}$ and $\mathbf{c}_{p}$, as $1 /|\mathcal{P}|\cdot\textstyle{\sum_{p\in\mathcal{P}^{rgb}}}\left |  \mathbf{M}^{v}_p(\mathbf{I}^{v}_p - \mathbf{c}_{p}(v) )\right |$.
We further constrain the silhouette errors by a mask loss ${L}_{mask}$ formed as $1 / (\alpha |\mathcal{P}|) \cdot \textstyle{\sum_{p\in\mathcal{P} \setminus \mathcal{P}^{rgb}}}(1-\mathbf{M}^{v}_p)\mathrm {CE}(\mathbf{M}^{v}_p, \mathbf{s} _{v,\alpha }(p))$,
where $\alpha$ is a hyperparameter, $\mathrm {CE}$ is a binary cross-entropy loss and $\mathbf{s}_{v,\alpha }(p)$ denotes predicted silhouette, written as $sigmoid(-\alpha \min_{t\geq 0} \mathcal{F}_{\theta }(\mathbf{x}_{p}))$. 
To enforce $\mathcal{F}_{\theta }$ to be a signed distance function,
we use the adaptive Eikonal regularization~\cite{gropp2020igr} $\mathcal{L}_{Eikonal}$ as $\mathbb{E}_{\mathbf{x}}\left ( \left \| \nabla_{\mathbf{x}} \mathcal{F}_{\theta}(\mathbf{x}) \right \| -1 \right ) ^{2}$.

Furthermore, we propose a registration loss to regularize the details of the implicit surface with the coarse mesh $\mathbf{G}_{c}$, such as the one produced by a 3DMM.
Similarly to \cite{NeuralPull}, our registration loss constrains the distance between $\mathbf{x}_{p}$ and its nearest coarse-mesh neighbor $\mathbf{x}^{\ast\dagger}_{p}$. 
In this way, geometric details are better preserved when the implicit surface changes greatly.
We assume a ray $\mathbf{r}$ has an intersection $\mathbf{x}_{p}$ with $\mathcal{G}$, and another one $\mathbf{x}^{\ast}_{p}$ with $\mathbf{G}_{c}$. 
Giving $\mathrm{x}^{\ast}_{p}$ to $\mathcal{F}_{\theta}$ as input,
we query the nearest neighbor $\mathrm{x}^{\ast\dagger}_{p}$ on the implicit surface by using its SDF values and specular normals, written as follows:
\begin{equation}
\mathbf{x}^{\ast\dagger}_{p} = \mathbf{x}^{\ast}_{p} - \left | \mathcal{F}_{\theta}\left ( \mathbf{x}^{\ast}_{p} \right )  \right | \cdot  \frac{\nabla \mathcal{F}_{\theta}\left ( \mathbf{x}^{\ast}_{p} \right )}{\left \| \nabla \mathcal{F}_{\theta}\left ( \mathbf{x}^{\ast}_{p} \right ) \right \|_{2} }.
\end{equation}
Therefore, the registration loss $\mathcal{L}_{registration}$ is defined as:
\begin{equation}
    \mathcal{L}_{registration} = \frac{1}{\mathcal{P}}\sum_{p\in\mathcal{P}^{rgb}} \left \| \mathbf{x}_{p} -  \mathbf{x}^{\ast\dagger } _{p} \right \|_{2}.
    \label{Eq_L_reg}
\end{equation}

Lastly, we devise a normal loss based on PBIDR (Sec.~\ref{Rendering3.2.2}) to regularize the cosine similarity between detailed specular normal $\mathbf{n}_{p}$ and diffuse normal $\mathbf{n}^{\ast\dagger } _{p}$. The normal loss $\mathcal{L}_{normal}$ can be written as:
\begin{equation}
    \mathcal{L}_{normal} = \frac{1}{\mathcal{P}}\sum_{p\in\mathcal{P}^{rgb}}(1-\dfrac{\mathbf{n}_{p} \cdot \mathbf{n}^{\ast\dagger } _{p}}{\max(\Vert \mathbf{n}_{p} \Vert _2 \cdot \Vert \mathbf{n}^{\ast\dagger } _{p} \Vert _2, 1e^{-8})}).
    \label{Eq_L_nor}
\end{equation}

Finally, the loss used for the optimization is defined as:
\begin{equation}
  \mathcal{L}_{total} =  \mathcal{L}_{photo} + \eta_{1} \mathcal{L}_{mask} + \eta_{2}\mathcal{L}_{Eikonal} + \eta_{3}\mathcal{L}_{registration} + \eta_{4}\mathcal{L}_{normal}.
\end{equation}
where $\eta_{1},\eta_{2},\eta_{3},\eta_{4}$ are all hyper-parameters.
\vspace{-6mm}
\subsubsection{Transfer Detailed SDF to Template Mesh}\label{Transfer3.2.4}
After optimizing $\mathcal{F}_{\theta}$, we can estimate any point's gradient and SDF value and ``pull'' \cite{NeuralPull} them to the implicit surface.
Therefore, we compute the SDF value for vertices in coarse 3D mesh and then approximate its normals. Then, we preserve the original mesh faces and give each vertex a displacement by its normal and SDF value, obtaining the fine-grained mesh $\mathbf{G}_{f}$ as follows:  
\begin{equation}
\mathbf{G}_{f} = \mathbf{G}_{c} + \sum_{p \in \mathcal{V}} n_{\mathcal{V}}\left (p  \right ) \mathcal{F}_{\theta}\left ( p \right ) ,
\end{equation}
where $\mathcal{V}$ is the set of all vertices in coarse mesh $\mathbf{G}_{c}$. 

\section{Implementation Details}
For the texture completion module (Sec.~\ref{texture3.1}),
we follow the same way as \cite{gecer2021ostec} to initialize latent parameters,
which accelerates the convergence and assists the optimizer to avoid local minima.
Then, the latent parameters (Eq.~\ref{eq1}), are optimized using an Adam \cite{kingma2014adam} optimizer with a learning rate of $0.25$.
Please note that our process (e.g., stitching, mask prediction~\cite{lin2021roi}) is fully automatic and does not require parameter tweaking for different images.

For the texture-guided 3D detail recovery (Sec.~\ref{Reconstruction3.2}), we are optimizing the $\mathcal{F}_{\theta}$, $\mathcal{A}$, $\mathcal{D}$ and $\mathcal{S}$ networks. 
$\mathcal{F}_{\theta}$ is an 8-layer MLP, including a skip-connection from input to the fourth layer, similar to \cite{yariv2020idr,ramon2021h3dnet}.
It uses Softplus as the activation function in all layers except for the output layer.
For the SDF, we use the sphere initialization technique from \cite{atzmon2020sal}, where the SDF initializes a unit sphere to stabilize the subsequent training process. Also, we utilize Position Encoding \cite{mildenhall2020nerf} to facilitate learning high-frequency information on the network.
Regarding the rendering networks, both the diffuse shading network $\mathcal{D}$ and specular shading network $\mathcal{S}$ use 1-layer MLP with ReLU activation function and an output layer with $\tanh$, 
while the diffuse albedo network $\mathcal{A}$ contains 4 layers.
We use Adam as the optimizer to train 400 epochs with a learning rate of $1e^{-4}$.
Our experiments for texture completion and details recovery have been performed on a single Nvidia V100 GPU in 1.2 hours total.
Please also note that our optimization framework keeps hyper-parameters the same for all subjects.

\section{Experiments}
In this section, we evaluate our approach in terms of texture and shape.
First, we assess the performance of our texture de-occlusion module and compare it with recent state-of-the-art completion algorithms under different inputs.
Then, we compare our detailed geometry recovery method with recent face reconstruction methods on NoW~\cite{sanyal2019learning}, 3DFAW~\cite{pillai20192nd}, and MICC Florence~\cite{bagdanov2011florence} dataset, respectively.
As the public face reconstruction datasets only contains very few cases
with mesoscopic frequency details, 
we manually select test cases with visible wrinkles,
in order to highlight the advantage of our algorithm in detailed shape recovery.
Furthermore, we show that our method can decompose the appearance of the inputs in a self-supervised way while reconstructing geometric details.

\subsection{Qualitative Results}\label{Sec.Qual.Results}

\begin{figure}[t]
\begin{floatrow}
\ffigbox[\FBwidth]
{\includegraphics[width=0.75\linewidth]{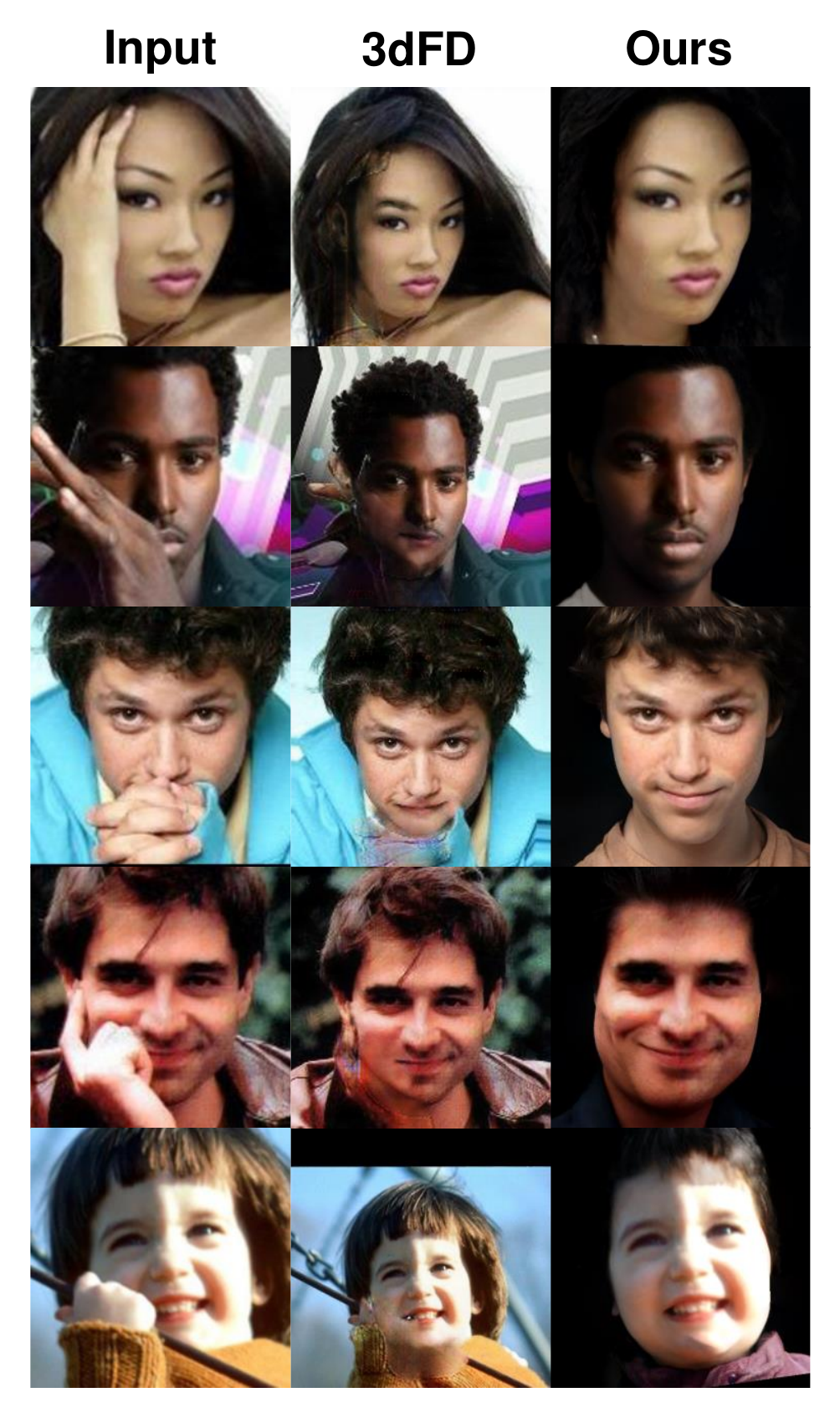}}
{\caption{\small Our de-occlusion results compared to 3dFD\cite{yuan2019face}. Input images and 3dFD results are also borrowed from~\cite{yuan2019face} for a fair comparison.}
\vspace{-0.4cm}
\label{fig:3dFD}}
\ffigbox[\FBwidth]
{\includegraphics[width=\linewidth]{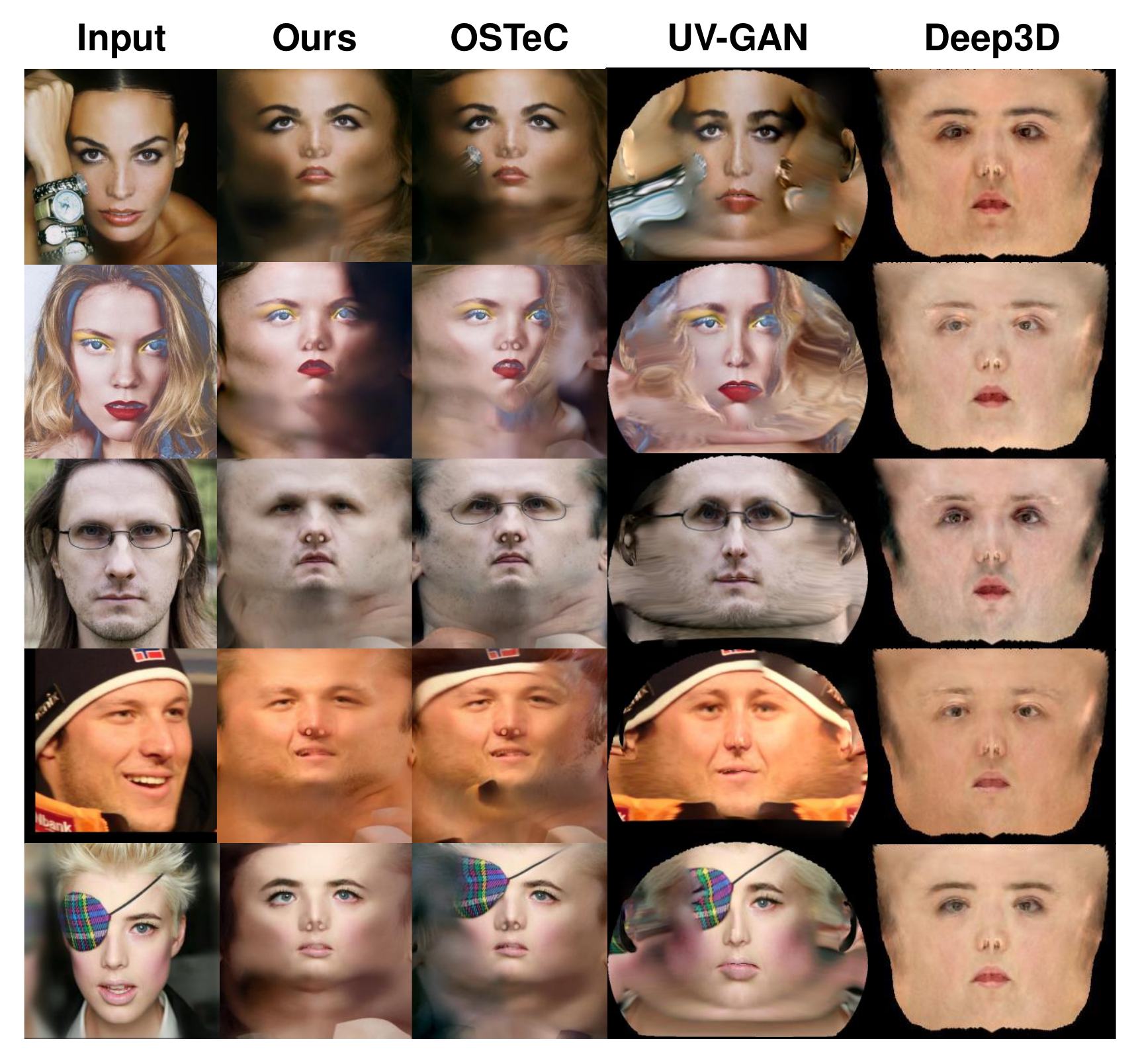}}
{\caption{\small Our texture-completion and de-occlusion results compared to OSTeC \cite{gecer2021ostec}, UVGAN \cite{deng2018uv} and Deep3D \cite{deng2019accurate}. Inputs are from the CelebAMask-HQ~\cite{CelebAMask-HQ} dataset.}
 \vspace{-0.4cm}
\label{fig:TextureCompletion}}
\end{floatrow}
\end{figure}

\noindent{\bf Face De-occlusion.}~Fig.~\ref{fig:3dFD} shows a comparison between our de-occlusion module and the SOTA method 3dFD \cite{yuan2019face}. 
The test images are borrowed from 3dFD's publicly released results. We observe that 3dFD always produces blurry results with arbitrary artifacts, while our method produces sharp and high-fidelity results.
Note that, only the facial area is in the scope of our method.

We further compare the full texture completion algorithm with Deep3d \cite{deng2019accurate}, UVGAN \cite{deng2018uv} and OSTeC \cite{gecer2021ostec}, as shown in Fig.~\ref{fig:TextureCompletion}. 
Deep3d \cite{deng2019accurate} employs a low-dimensional parametric texture space and which generates unrealistic results.
UVGAN \cite{deng2018uv} and OSTeC \cite{gecer2021ostec} produce consistent results,
but are suffering from the inability to handle occlusions. 
As our method includes a robust de-occlusion module, our results significantly outperform the texture completion results of counterparts.

\begin{figure*}[t]
    \centering
    \includegraphics[width=\linewidth]{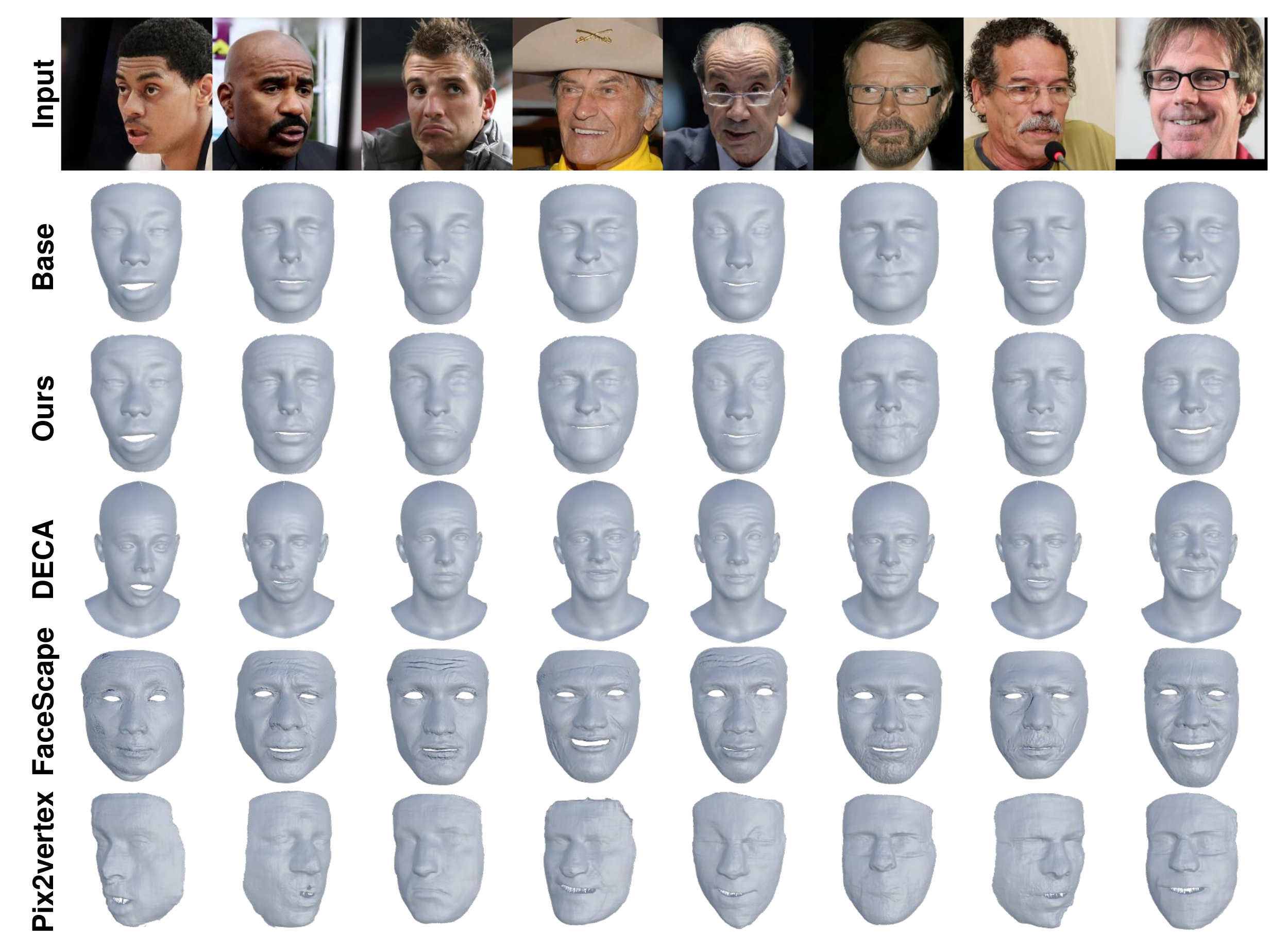}
    \vspace{-0.2cm}
    \caption{\small Comparisons to single-view detailed face reconstruction methods, from top to bottom: Inputs, Base~\cite{gecer2019ganfit}, Ours, DECA \cite{feng2021learning} (trained on 2M images), FaceScape \cite{yang2020facescape} (trained on 3D scans) and Pix2Vertex \cite{sela2017unrestricted} (synthetic dataset).}
     \vspace{-0.2cm}
    \label{fig:Figure5_single_view_mesh}
\end{figure*}

\noindent{\bf Details Recovery.}~For the detail recovery part, 
we compare with recent single-image reconstruction methods that reconstruct mesoscopic details, such as wrinkles, namely 
DECA \cite{feng2021learning}, FaceScape \cite{yang2020facescape} and Pix2Vertex \cite{sela2017unrestricted}.
As shown in Fig.~\ref{fig:Figure5_single_view_mesh},
our reconstruction produces accurate details,
even in examples with significant contouring, wrinkles, or occlusions.
Pix2Vertex \cite{sela2017unrestricted} produces the smooth results, 
omitting wrinkles and failing under occlusions.
DECA \cite{feng2021learning} can generate wrinkles robustly,
but the generated details always exhibit a similar pattern.
Arguably, FaceScape \cite{yang2020facescape} reconstructs accurate details,
but fails in occluded images and omits details in some examples.
Moreover, please note that DECA trains on 2 million images \cite{feng2021learning}, and FaceScape is trained on high-fidelity 3D scans and high-quality displacement maps.
On the contrary, we achieve comparable results by using only one single image and an off-the-self 3DMM as initialization.

\begin{figure*}[t]
    \centering
    \includegraphics[width=\linewidth]{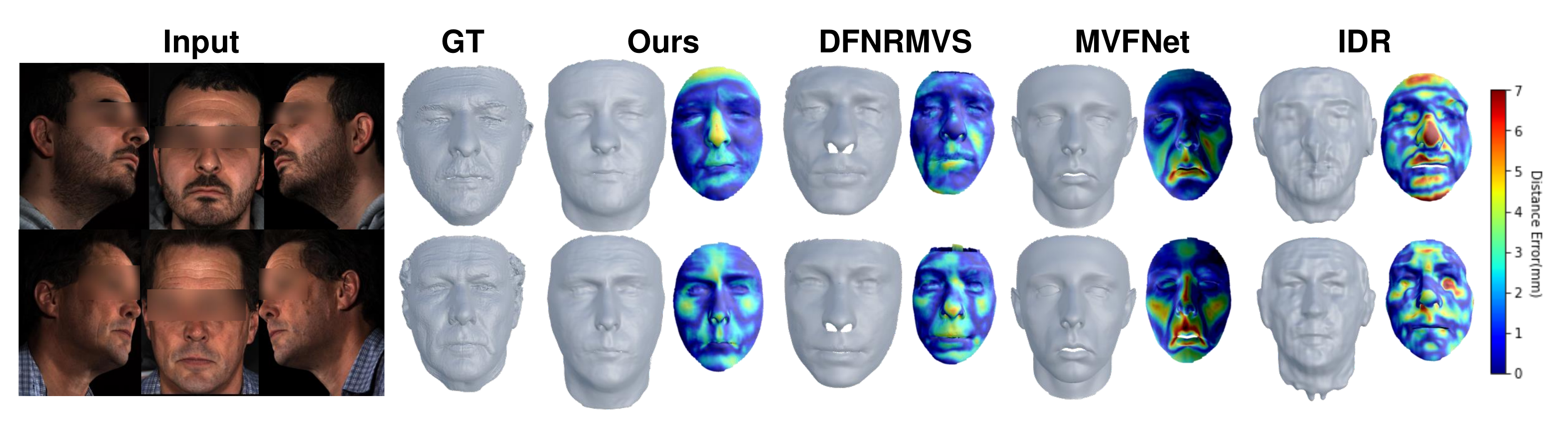}
    \vspace{-0.4cm}
    \caption{\small Comparisons of our method with multi-view reconstruction methods, e.g., DFNRMVS \cite{bai2020deep}, MVFNet \cite{wu2019mvf} and IDR \cite{yariv2020idr}. Our method only takes one frontal view as the input while others utilize all three views. Input and ground truth (GT) are described in Sec.~\ref{Sec.Qual.Results}}.
    \vspace{-0.4cm}
    \label{fig:Figure6_multi_view_mesh}
\end{figure*}

In addition, we also compare with multi-view reconstruction methods, 
including MVF-Net~\cite{wu2019mvf}, DFNRMVS~\cite{bai2020deep} and IDR~\cite{yariv2020idr}. 
These three methods require three views, while ours is single-view.
To evaluate these multi-view based methods qualitatively, we collect one hundred 3D face scans with a coarse geometry reconstructed with SfM \cite{schoenberger2016sfm}.
Then we use a state-of-the-art network \cite{lattas2021avatarme++}
to acquire detailed normals from the scanned texture 
and emboss them to the coarse mesh \cite{nehab2005efficiently}, 
synthesizing detailed pseudo-ground truths (GT) with realistic details, as shown in Fig. \ref{fig:Figure6_multi_view_mesh}.
MVF-Net \cite{wu2019mvf} and DFNRMVS \cite{bai2020deep}, which are based on the 3DMM model, obtain plausible meshes with similar contour but fail to capture
medium-frequency details such as forehead wrinkles and eyelids.
Most similar to our method, IDR \cite{yariv2020idr} is limited by the number of input views, and reconstructs noisy meshes.
On the contrary, our method only uses a single image as input but creates auxiliary views by recovered textures map, which helps recover realistic shapes with accurate details and wrinkles.

As a bonus, our PBIDR decomposes the rendered texture into approximate diffuse albedo, diffuse shading, and specular shading in a self-supervised way, as shown in Fig. \ref{fig:Figure7_decomposition}. As our decomposition is totally self-supervised, diffuse albedo is sub-optimal.
Nevertheless, these components are decoupled without any training data and can be used to re-render the reconstruction with different shading, especially under complex lighting conditions (as shown in our supplemental materials).

\subsection{Quantitative Results}
\noindent{\bf Face De-occlusion.}~In order to quantitatively evaluate our face de-occlusion method,
we measure Peak Signal-to-Noise Ratio (PSNR),
Structural Similarity Index (SSIM),
Perceptual Similarity (LPIPS),
and Fréchet Inception Distance (FID),
on images from the 3dFD \cite{yuan2019face} test data.
Tab.~\ref{table:DeOcclusion} shows that our method significantly outperforms 3dFD.

\begin{table}[t]
\begin{floatrow}
\capbtabbox{
\setlength{\tabcolsep}{4pt}
\resizebox{0.4\textwidth}{!}{\begin{tabular}{@{}lllll@{}}
\toprule
~Methods & \multicolumn{1}{c}{PSNR$\uparrow$} & \multicolumn{1}{c}{SSIM$\uparrow$} &  LPIPS$\downarrow$ & FID$\downarrow$ ~ \\ \midrule
~3dFD \cite{yuan2019face}   &           11.71               &            0.3946              & 0.2543         &   267.87~      \\ \midrule
~\textbf{Ours}   &           \textbf{15.83}               &            \textbf{0.5481}              &     \textbf{0.1359}    &   \textbf{208.08} ~     \\ \bottomrule
\end{tabular}}}
{\caption{\small  Comparison with 3dFD \cite{yuan2019face} for the face de-occlusion on test data provided by 3dFD.}
 \label{table:DeOcclusion}
}
\capbtabbox{
\setlength{\tabcolsep}{4pt}
 \resizebox{0.5\textwidth}{!}{
 
\begin{tabular}{@{}lllll@{}}
\toprule
~Methods       & w.O PSNR     & w.O SSIM      & wo.O PSNR & wo.O SSIM~ \\ \midrule
~UVGAN~\cite{deng2018uv}         & 19.36          & 0.8413          & 21.84       & 0.8945~      \\
~OSTeC~\cite{gecer2021ostec}         & 19.91          & 0.8549          & 22.34       & 0.9035 ~     \\ \midrule
~\textbf{Ours} & \textbf{22.08} & \textbf{0.8973} & 22.32       & 0.9018   ~   \\ \bottomrule
\end{tabular}
}}
{
 \caption{\small Evaluations of occlusion-robust UV completion on the Multi-PIE~\cite{gross2010multipie} dataset. Note that we manually add occlusions for evaluation.}
 \label{table:TextureCompletion}
}
\end{floatrow}
\end{table}

Furthermore, we compare our robust texture completion methods quantitatively with previous OSTeC~\cite{gecer2021ostec}, UVGAN~\cite{deng2018uv}, and Deep3D~\cite{deng2019accurate} on the UVDB (Multi-PIE~\cite{gross2010multipie}) dataset. 
We randomly select 100 subjects, and half of them have been manually added occlusions as this dataset hardly contains occlusion cases.
Then, PSNR and SSIM are employed between the estimated UV maps and the ground truth.
Our approach achieves comparable results with OSTeC~\cite{gecer2021ostec} on the non-occluded cases and significantly outperforms all methods on the occluded examples, shown in Tab.~\ref{table:TextureCompletion}.

\noindent{\bf Details Recovery.}~Since there is no benchmark for comparing detailed face shape recovery, we first compare ours with DECA \cite{feng2021learning}, the state-of-the-art detailed single face reconstruction approach, on the NoW \cite{sanyal2019learning} validation dataset.
In Tab.~\ref{table:single-view}, both ours and DECA~\cite{feng2021learning} heavily rely on the reconstructed coarse mesh, while details hardly contribute to absolute distance error.

Besides, we conduct a numerical comparison with MVFNet \cite{wu2019mvf}, DFNRMVS \cite{bai2020deep} and IDR \cite{yariv2020idr} 
on 3DFAW \cite{pillai20192nd}, MICC Florence \cite{bagdanov2011florence} and our scans, as shown in Tab.~\ref{table:Multi-view} and Fig.~\ref{fig:Figure6_multi_view_mesh}. 
We register the GT and reconstructed meshes to the same template, perform rigid ICP alignment, and crop the tight face part,
which includes details such as wrinkles and is registered without errors.
The evaluation indicators are the same as those of MICC \cite{bagdanov2011florence}.
Again, the absence of details in the datasets \cite{pillai20192nd,bagdanov2011florence} inadvertently increases the error when evaluating our detailed reconstructions.
In addition, we compare their normal vector cosine distance errors on our collected dataset.
The results in Tab.~\ref{table:Multi-view} exhibit that our approach achieves the best performance and is more accurate in normal vectors.

\begin{table}[t]
\begin{floatrow}
\capbtabbox{
\setlength{\tabcolsep}{4pt}
\resizebox{0.35\textwidth}{!}{\begin{tabular}{@{}lccc@{}}
\toprule
\multicolumn{1}{c}{Methods}                         & Median & Mean & Std ~\\ \midrule
~DECA coarse \cite{feng2021learning}                & 1.18       & 1.46     & 1.25 ~\\ 
~DECA detail \cite{feng2021learning}                & 1.19       & 1.47     & 1.15 ~\\ \midrule
~Ours base  \cite{gecer2019ganfit}                  & 1.23       & 1.52     & 1.31 ~\\ 
~Ours detail                                        & 1.23       & 1.53     & 1.31 ~\\ \bottomrule
\end{tabular}}}
{\caption{\small Comparison with DECA \cite{feng2021learning} on the NoW \cite{sanyal2019learning} validation dataset.}
 \label{table:single-view}
}
\capbtabbox{
\setlength{\tabcolsep}{4pt}
 \resizebox{0.52\textwidth}{!}{
\begin{tabular}{@{}lcccc@{}}
\toprule
\multicolumn{1}{c}{Method} & 3DFAW     & MICC      & UC. Distance & UC. Normal~ \\ \midrule
~MVFNet~\cite{wu2019mvf}    & 2.27$\pm$0.54 & 1.34$\pm$0.29 &  1.53$\pm$0.56   & 0.094 ~\\
~DFNRMVS~\cite{bai2020deep} & 2.26$\pm$0.53 & 1.32$\pm$0.28 &  1.15$\pm$0.41   & 0.087 ~\\
~IDR~\cite{yariv2020idr}    & 6.48$\pm$2.83 & 4.18$\pm$1.35 &  2.84$\pm$1.32   & 0.131 ~\\ \midrule
~Base~\cite{gecer2019ganfit} & 2.06$\pm$0.44 & 1.23$\pm$0.22 & 1.22$\pm$0.45   & 0.088 ~\\
~Ours                       & 2.02$\pm$0.42 & 1.19$\pm$0.21 &  \textbf{1.20$\pm$0.43}   & \textbf{0.069} ~\\ \bottomrule
\end{tabular}
}}
{
 \caption{\small Reconstruction error (mm) on the 3DFAW \cite{pillai20192nd}, MICC \cite{bagdanov2011florence} and our under-control (UC) scans and normal cosine distance on ours.}
 \label{table:Multi-view}
}
\end{floatrow}
\end{table}

\subsection{Ablation Study}
\begin{figure}[t]
\begin{floatrow}
\ffigbox[\FBwidth]
{\includegraphics[width=\linewidth]{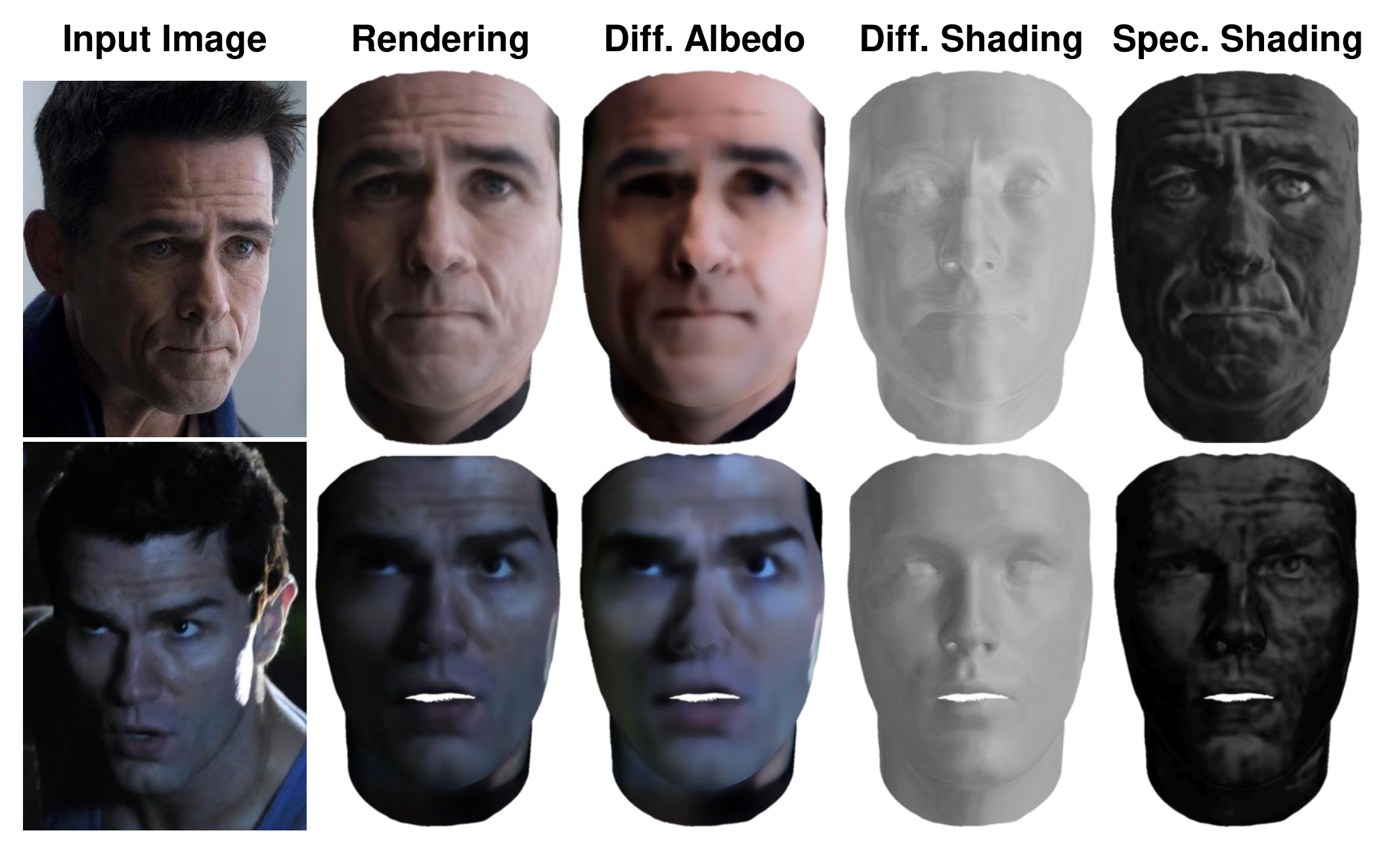}}
{\caption{\small Self-supervised decomposition. Left to right: input, re-rendering, approximate diffuse albedo, diffuse shading and specular shading.}
\label{fig:Figure7_decomposition}}
\ffigbox[\FBwidth]
{\includegraphics[width=0.9\linewidth]{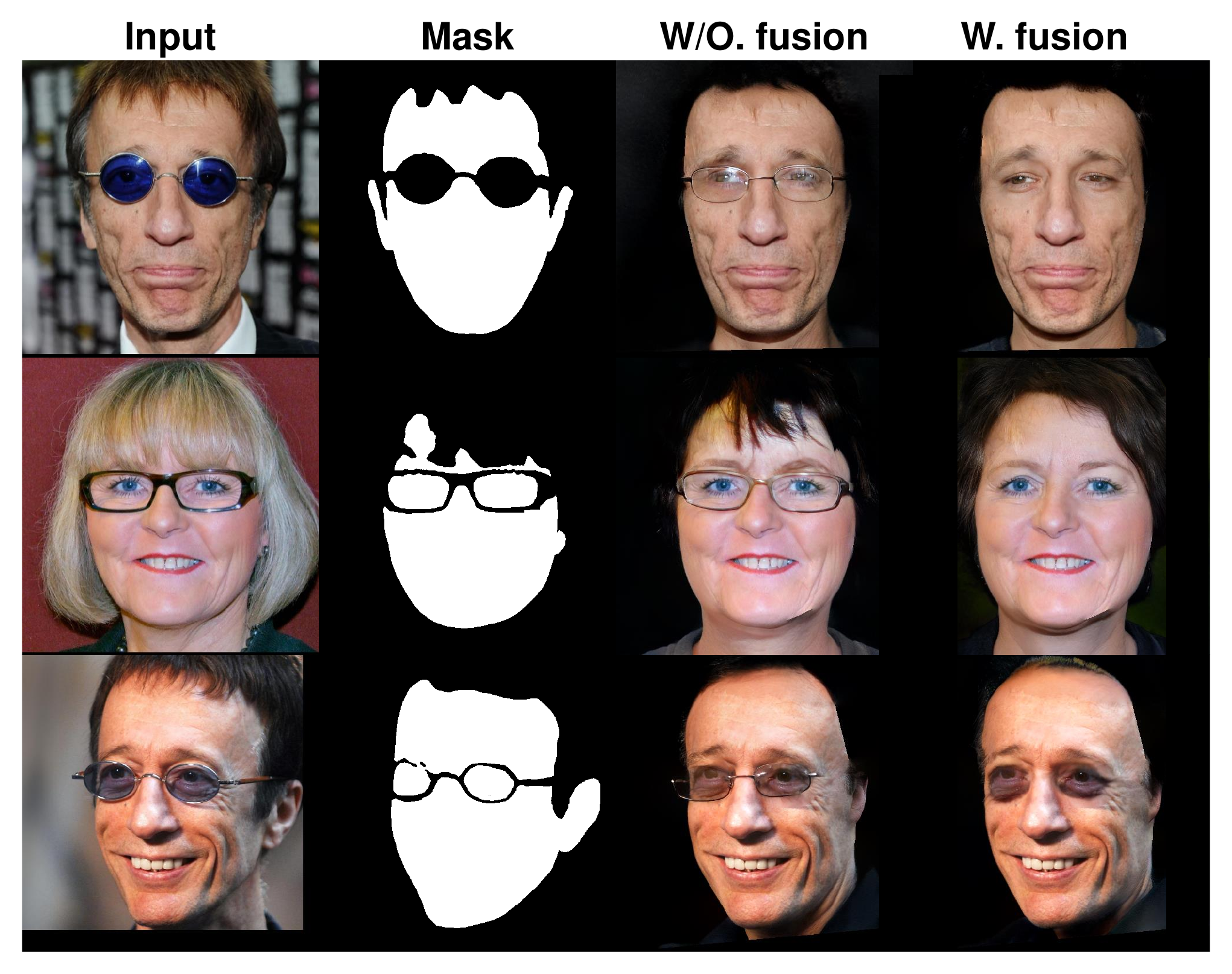}}
{\caption{\small Ablation study of our self-fusion. Left to right: inputs, facial masks, without and with our module.}
\label{fig:Figure8_ablation_deocclusion}}
\end{floatrow}
\vspace{-1mm}
\end{figure}

\begin{figure}[t]
\begin{floatrow}
\ffigbox[\FBwidth]
{\includegraphics[width=0.9\linewidth]{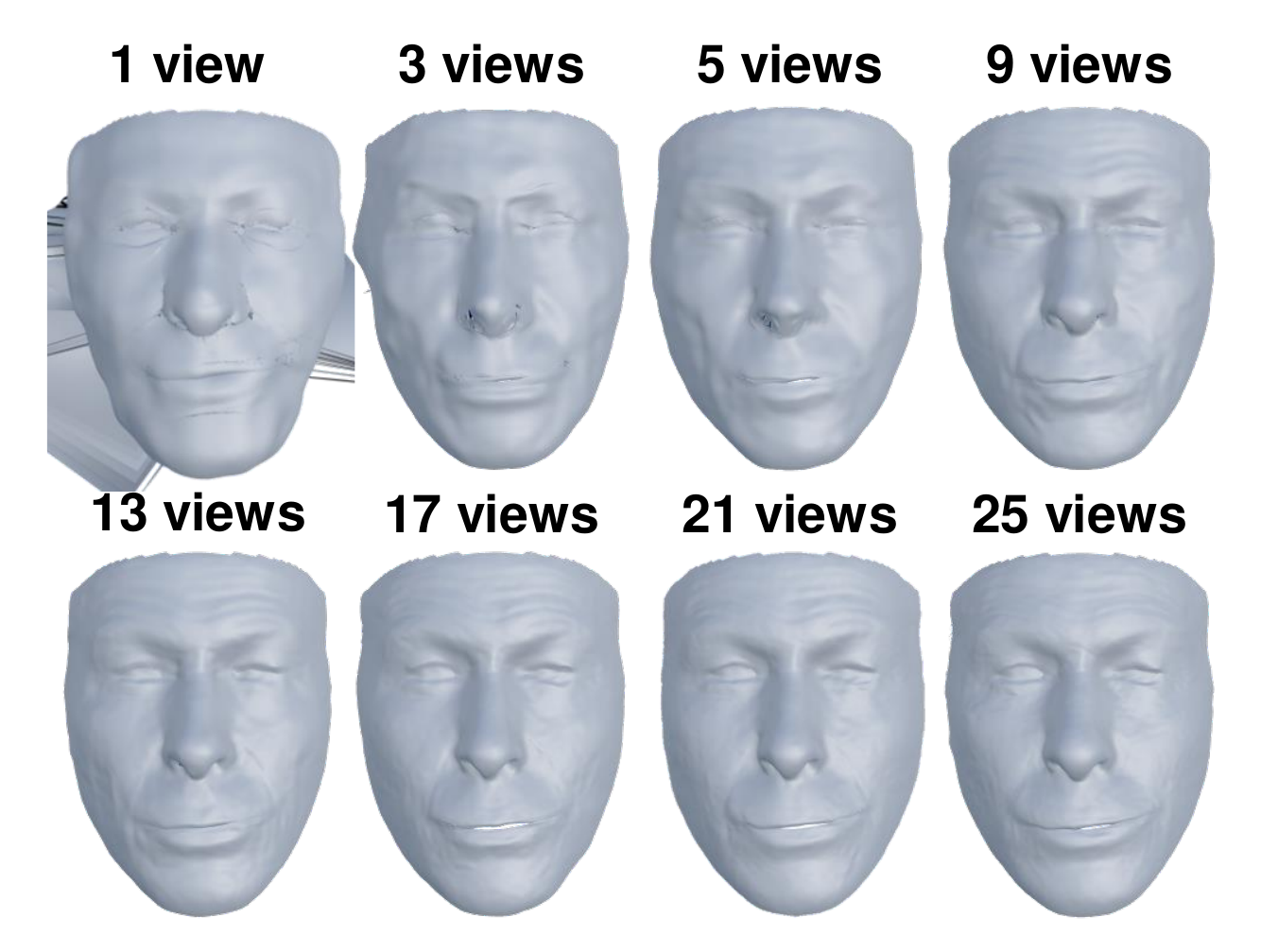}}
{\caption{\small Ablation study of rendered auxiliary views. Using only a single view will cause depth ambiguity.}
\label{fig:Figure9_ablation_views}}
\ffigbox[\FBwidth]
{\includegraphics[width=\linewidth]{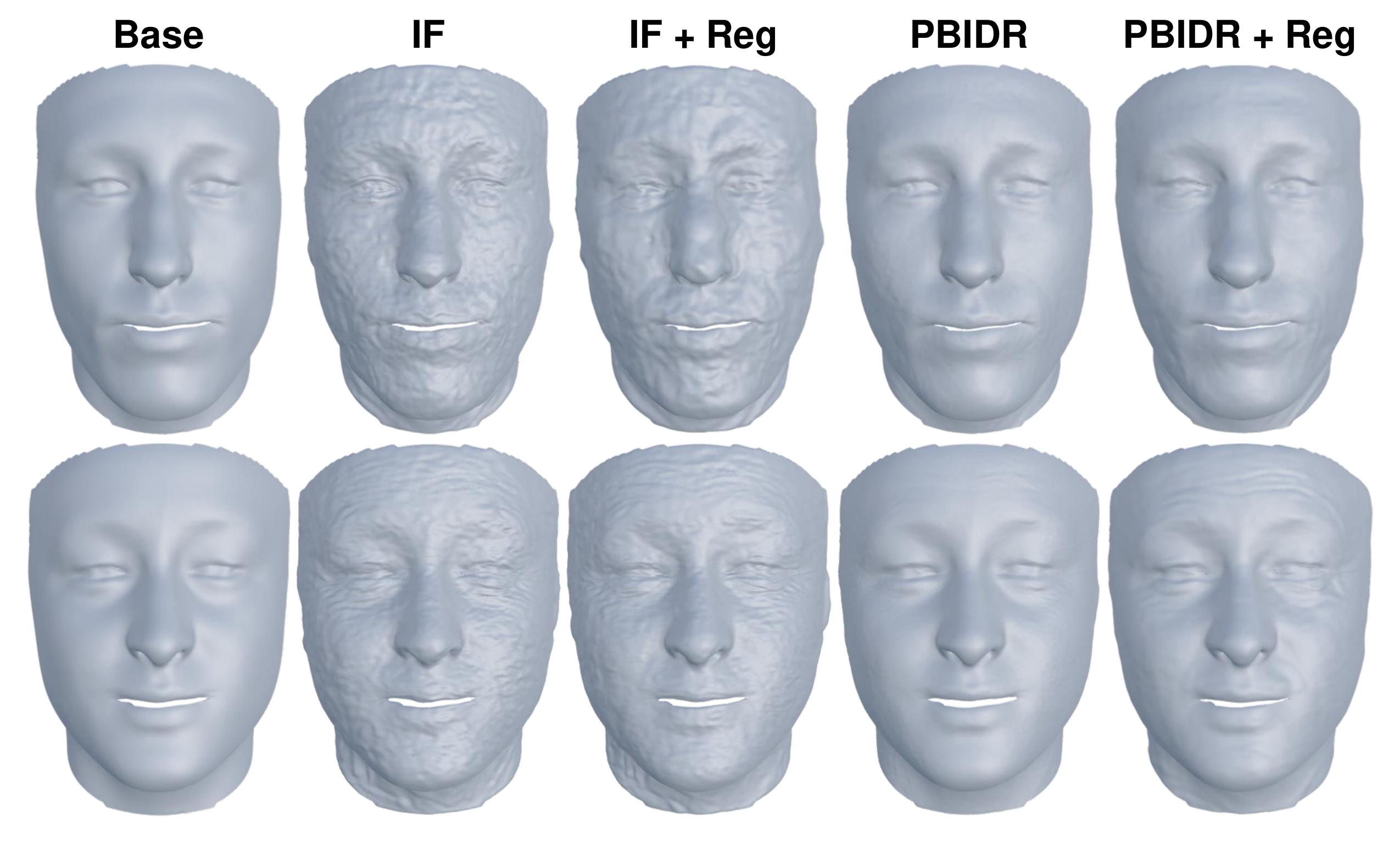}}
{\caption{\small Ablation study of our pipeline, where ``IF'' means Implicit Function and ``Reg'' is the registration loss.}
\label{fig:Figure10_ablation_pbidr}}
\end{floatrow}
\vspace{-1mm}
\end{figure}

\noindent{\bf Self Fusion Module.}
First, we validate our self-fusion module for the texture completion.
We expect the generated image to be reasonably extended based on the reference image. As seen from Fig.~\ref{fig:Figure8_ablation_deocclusion}, our self-fusion module enables the generator to focus on the face without reproducing external occlusions.
In addition, the last row of Fig.~\ref{fig:Figure8_ablation_deocclusion} shows a failure case of our method on the eyes, because of the incorrect segmentation mask.

\noindent{\bf Auxiliary Images Set.}~Next, we evaluate the number of rendered views in the auxiliary images set.
We set the maximum yaw angle as $45^{\circ}$  to prevent the ray from casting the back of the face manifold as our face is not watertight.
Then, we render different numbers of views ( 1, 3, 5, 9, 13, 17, 21, and 25 images), evenly sampling from $-45^{\circ}$ to $+45^{\circ}$.
As seen from Fig.~\ref{fig:Figure9_ablation_views}, using only a single view causes depth ambiguity. The quality of reconstructed details increases as the quantity of views increases. When the quantity of views exceeds 20, reconstructed details are saturated. We choose 17 views to balance quality and time consumption.

\noindent{\bf Detail Recovery Pipeline.}~We qualitatively study the impact of PBIDR and registration loss in our method. As shown in Fig.~\ref{fig:Figure10_ablation_pbidr}, directly using implicit functions will introduce a lot of high-frequency noise and artifacts.
Our PBIDR module significantly reduces noise artifacts while using implicit function as detail representation.
In addition, our registration loss makes the embossing details more prominent and better preserves the details in the implicit surface.
Finally, our approach achieves accurate detail recovery under the combination of PBIDR and registration loss, and rarely generates additional noise and artifacts.

\section{Conclusions}

In this paper, 
we propose a novel facial geometric detail recovery algorithm by using only one single face image. Our method is divided into two steps, the first restoring completed high-fidelity texture and the second using a physically-based implicit differentiable renderer to enhance the details of the coarse facial geometry. 
Extensive experiments show that our algorithm achieves comparable shape reconstruction quality and the best detail recovery, especially for single image input.
Despite using implicit functions, our implicit-to-mesh transfer method ensures that our detailed geometry can be used in applications that require triangular meshes.

Since our method is self-supervised, it is not limited by imbalanced datasets on race and age. However, the proposed method has the following limitations.
(1) Our method utilizes off-the-shelf 3DMM fitting results as a self-supervised label, which leads to reconstruction results relying on the fitting results.
(2) Imprecise segmentation will introduce artifacts in the de-occlusion results, shown in Fig.~\ref{fig:Figure8_ablation_deocclusion}.
(3) Our method is optimization-based and it is uneconomical in time. In the future, we will try to decrease the toolkit dependency and accelerate our method.

\noindent\textbf{Acknowledgements.}
This work was supported in part by NSFC (61906119, U19B2035), Shanghai Municipal Science and Technology Major Project\\ (2021SHZDZX0102).
Stefanos Zafeiriou acknowledges support from the EPSRC Fellowship DEFORM (EP/S010203/1), FACER2VM (EP/N007743/1) and a Google Faculty Fellowship.

\bibliographystyle{splncs04}
\bibliography{arxiv}
\clearpage
\appendix

\section*{Supplemental Materials}

In this supplemental materials document,
we first show a detailed algorithm that describes our method (Sec.~\ref{sec1_algorithm}) and discuss the required computational costs (Sec.~\ref{sec1_computation}),
and describe the datasets used for our comparisons (Sec.~\ref{sec2}).
Then we show additional applications results,
such as relighting (Sec.~\ref{sec3}) and expression manipulation (Sec.~\ref{sec4}). Finally, we show additional texture completion results
in Sec.~\ref{sec5}.

Moreover, \textbf{we attach a video} with high-resolution rotating rendered mesh results, which showcases the quality and accuracy of the texture completion and geometric details acquisition.
We also attach \textbf{the source code} of our PBIDR module and a pre-trained model.
We will make \textbf{our code publicly available} before the publication date.

\section{Algorithm Overview}
\label{sec1_algorithm}
We summarize the texture completion (Sec.~3.1),
shape refinement (Sec.~3.3) and details registration to 3DMM (Sec.~3.4) steps of our novel method in Algorithm~\ref{Ours_al}. 

The algorithm uses the following notation, as described in the main paper:
$3DMM$ is a common 3D Morphable Model fitting method \cite{gecer2019ganfit},
$Segment$ is an off the-self facial segmentation model \cite{lin2021roi},
$FaceGen$ is a facial generator \cite{Karras2019stylegan2} (Sec.~3.1.1),
$OSTeC$ is the generator used by \cite{gecer2021ostec} (Sec.~3.1.2),
$\mathcal{L}_{total}^{T}$ is defined in Eq~(1) and
$Render$ is an off-the-shelf mesh renderer \cite{ravi2020pytorch3d} (Sec.~3.1.3).
Moreover,
$SphereTracing$ is an implicit sphere tracing function \cite{jiang2020sdfdiff} (Sec.~3.2.1),
$RayMarching$ is a differentiable ray marching function \cite{yariv2020idr},
$\mathcal{F}_\theta$ is the network encoding the SDF shape $SDF_p$ and feature vector $\mathbf{z_p}$,
$BaS$ denotes a function that calculates normals from a mesh barycentric coordinates (Sec.~3.2.2),
$\mathcal{A}$ is the diffuse albedo network of PBIDR,
$\mathcal{D}$ is the diffuse shading network of PBIDR, $\mathcal{S}$ is the specular shading network of PBIDR, $a_s$ denotes the constant specular albedo value and $\mathcal{L}_{total}^{G}$ is the PBIDR loss function defined in Eq~(8).

\begin{algorithm}[ht]
\caption{Facial Geometric Detail Recovery via Implicit Representation}
\label{Ours_al}
\LinesNumbered 
\KwIn{Input Face Image: $\mathbf{I}_{i}$}
\KwOut{Completed UV Texture Map: $\mathbf{T}_{o}$}
\KwOut{Detailed Mesh: $\mathbf{G}_{f}$}
$\mathbf{{G}_c, c} \gets 3DMM(\mathbf{I}_i)$\\
$\mathbf{M_i} \gets Segment(\mathbf{I}_i)$\\
\For{each step}{
$\mathbf{I}_{r} \gets FaceGen(\mathbf{I}_{i}, \mathbf{M}_{i})$\\
$\mathbf{I}_{f} \gets  \mathbf{M}_{i}\mathbf{I}_{i} + (1-\mathbf{M}_{i})\mathbf{I}_{r}$\\
Minimize 
$\mathcal{L}_{total}^{T}(\mathbf{I_i, I_r, I_f}$
}
$\mathbf{T}_{o} \gets OSTeC(\mathbf{I}_{f}, \mathbf{G}_{c}, \mathbf{c})$\\
$\mathbf{I}^{\ast} \gets Render(\mathbf{G}_{c}, \mathbf{T}_{o}, v=1,2,\dots,24)$\\
\For{each step}
{Sample a mini-batch of pixels $\mathcal{P}$ from view direction $\mathbf{v}$ and image $\mathbf{I}^{v}$\;
\For{each $\mathbf{p} \in \mathcal{P}$}{
$\mathbf{x}_{p} \gets SphereTracing(\mathbf{p}, \mathbf{v}, \mathcal{F}_{\theta})$\\
$\mathbf{x}^{\ast}_{p} \gets RayMarching(\mathbf{p}, \mathbf{v}, \mathbf{G}_{c})$\\
$\mathbf{SDF}_{p}, \mathbf{z}_{p} \gets \mathcal{F}_{\theta }(\mathbf{p})$\\
$\mathbf{n}_{p}, \mathbf{n}^{\ast}_{p} \gets \nabla_\mathbf{x}\mathcal{F}_{\theta}(\mathbf{x}_{p}), BaS(\mathbf{G}_{c}, \mathbf{x}^{\ast}_{p})$ \\
$\mathbf{c}_{p} ( \mathbf{v}) \gets \mathcal{A}(\mathbf{x}_{p}, \mathbf{z}_{p}) \mathcal{D}(\mathbf{n}^{\ast}_{p}) + a_s \; \mathcal{S}(\mathbf{n}_{p}, \mathbf{v})$\\
}
Minimize the loss $\mathcal{L}_{total}^{G}(\mathbf{x}_{p}, \mathbf{x}^{\ast}_{p}, \mathbf{c}_{p}, \mathcal{F}_{\theta})$\\
}
$\mathbf{G}_{f} \gets \mathbf{G}_{c} + \sum_{p \in \mathcal{V}} n_{\mathcal{V}}\left (p  \right ) \mathcal{F}_{\theta}\left ( p \right ) $ \\
\Return $\mathbf{T}_{o}$, $\mathbf{G}_{f}$
\end{algorithm}

\section{Computational Costs}
\label{sec1_computation}
Our method requires approximately 12~minutes reconstruct a completed UV (Sec.~3.1),
1~hour to reconstruct the refined (Sec~3.3) and registered (Sec.~3.4) mesh, 
and 5 minutes for single image rendering using PBIDR (Sec.~3.2). 
For each input image, our approach needs to optimize an implicit surface from scratch based on the generated high-resolution multi-view images,
which significantly increases the overall computational cost. 
In addition, the current step of ray-marching is CPU-intensive operation, whose efficiency could be significantly improved in future work. 
Although requiring a significant amount of time for a single reconstruction, we still have a great advantage in a single-shot reconstruction scenario, considering we do not need to collect additional data for training.

\section{Datasets}\label{Supp.data}
\label{sec2}

\noindent{\bf NoW Benchmark:} The NoW \cite{sanyal2019learning} Benchmark consists of a validation set (20 subjects) and a test set (80 subjects). In the validation set, each subject has a 3D face scan corresponding to a neutral expression.
Since we mainly evaluate the ability to reconstruct face details from a single photo, we only select neutral expressions with 3D ground truth as test cases.

\noindent{\bf 3DFAW dataset:} The 3DFAW \cite{pillai20192nd} dataset consists of 26 subjects with corresponding video and 3D ground truth mesh. For the finer evaluation in the face region, they cut both the reconstructions and the ground truth using a sphere of 95 mm radius and with center at the tip of the nose of the ground truth mesh and refine the alignment with rigid ICP. Finally, the Average Root Mean Square Error Metric (ARMSE) is calculated to evaluate the reconstructed mesh quality.

\noindent{\bf MICC Florence dataset:} The dataset \cite{bagdanov2011florence} provides 3D scans of 53 subjects and short video samples.
Moreover, GANFit \cite{gecer2019ganfit} adopts the same tight mask as 3DFAW and average symmetric point-to-plane distance as an evaluation indicator.

\noindent{\bf Our collected datasets:} We collect one hundred 3D face scans with a coarse geometry reconstructed with SfM \cite{schoenberger2016sfm}
Then we use a state-of-the-art network \cite{lattas2021avatarme++}
to acquire detailed normals from the scanned texture 
and emboss the normals to the coarse mesh \cite{nehab2005efficiently}, 
synthesizing detailed pseudo-ground truths (GT) with realistic details. We register the GT and reconstructed meshes to the same template, perform rigid ICP alignment, and crop the tight mask for facial detail comparison.

\section{Illumination Transfer and Relighting}
\label{sec3}
Our method can also be used for relighting application,
for example illumination transferring between two images,
or rendering under novel illumination conditions.

\begin{figure}[ht]
    \centering
    \includegraphics[width=0.96\linewidth]{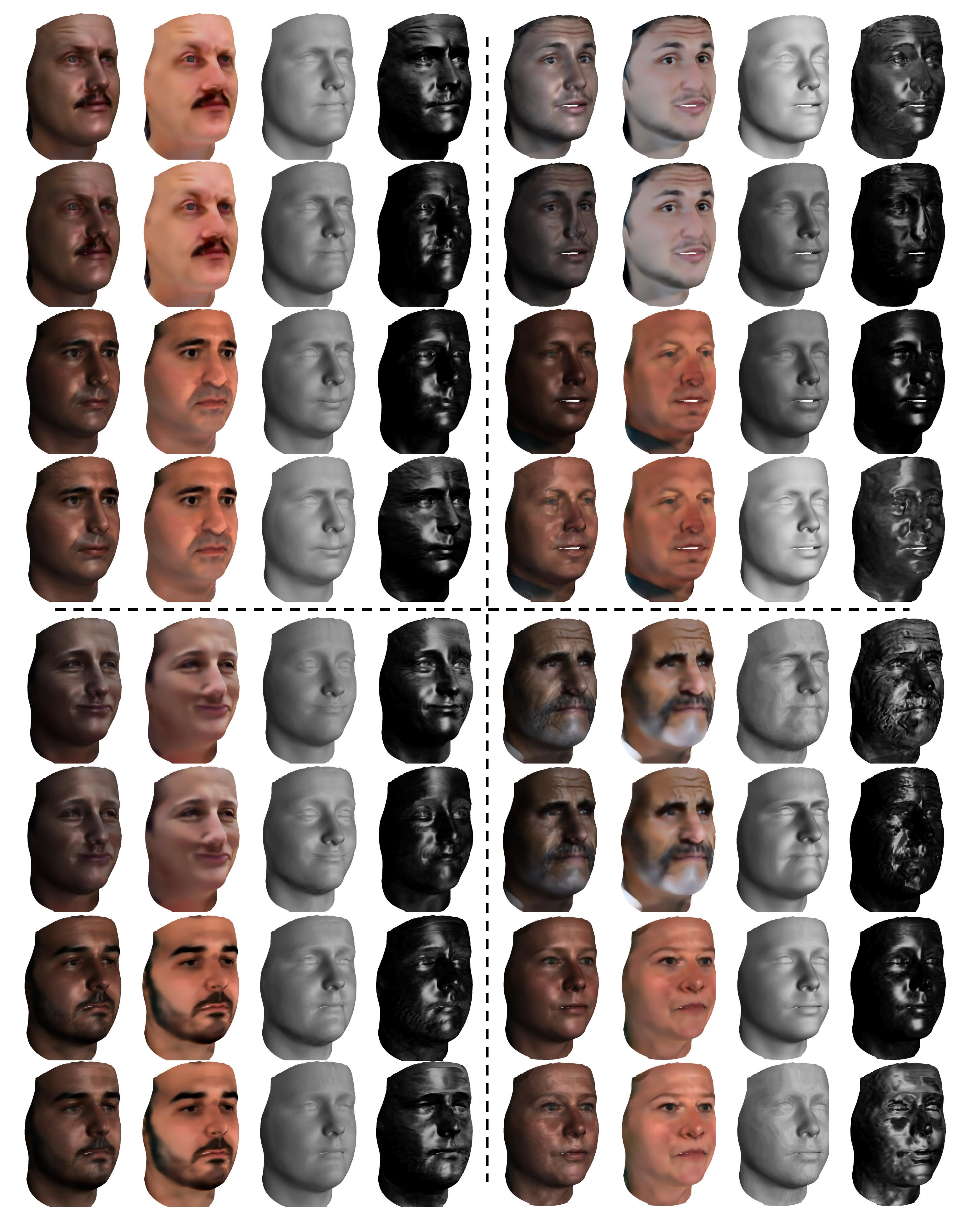}
    \caption{
    We decompose two images under different lighting conditions in an unsupervised way using PBIDR (Sec.~3.3)
    and exchange their diffuse $\mathcal{D}$ and specular shading $\mathcal{S}$ components,
    while keeping their diffuse albedo $\mathcal{A}$ and shape $G$.
    For each $4\times4$ square,
    each column shows the PBIDR-rendered subject,
    diffuse albedo $\mathcal{A}$, diffuse shading $\mathcal{D}$
    and specular shading $\mathcal{S}$;
    the first and third rows show the PBIDR results for the input image,
    while the second and fourth row show the swapped illumination for each subject.
    }
    \label{suppfig: relighting_transfer}
\end{figure}

\begin{figure}[ht]
    \centering
    \includegraphics[width=0.96\linewidth]{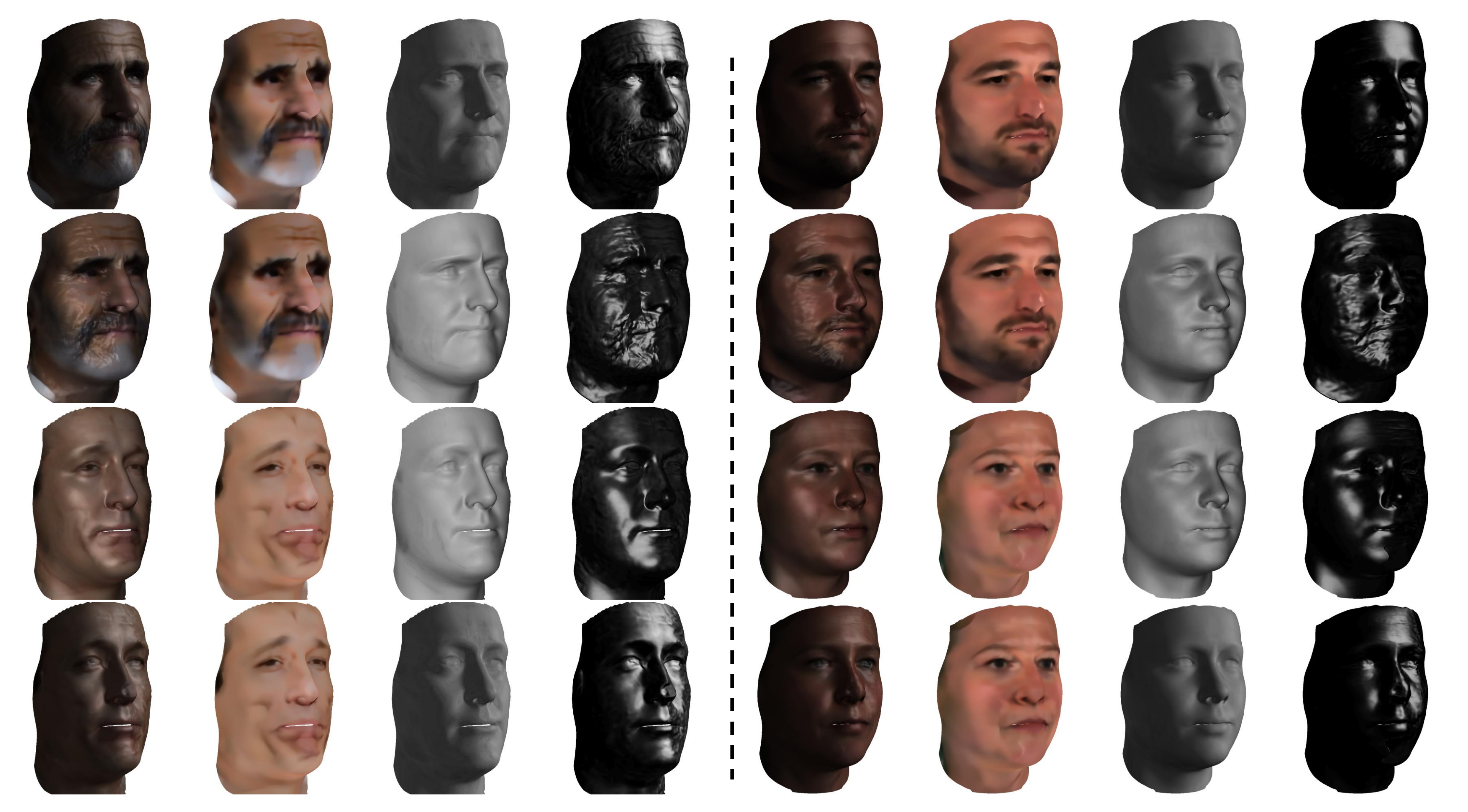}
    \caption{
    We introduce two significantly different light sources,
    with up-left and down-right directions
    and use these to render the depicted subjects.
    Then we decompose we exchange the illumination between the two states, an unsupervised way using PBIDR (Sec.~3.3),
    while keeping their diffuse albedo $\mathcal{A}$ and shape $G$.
    For each $4\times4$ square,
    each column shows the PBIDR-rendered subject,
    diffuse albedo $\mathcal{A}$, diffuse shading $\mathcal{D}$
    and specular shading $\mathcal{S}$;
    the first and third rows show the PBIDR results for the input image,
    while the second and fourth row show the swapped illumination for each subject.
    }
    \label{suppfig: relighting_novel}
\end{figure}

In Fig.~\ref{suppfig: relighting_transfer}, we decompose two images under different lighting conditions in an unsupervised way using PBIDR (Sec.~3.3)
and exchange their diffuse $\mathcal{D}$ and specular shading $\mathcal{S}$ components,
while keeping their diffuse albedo $\mathcal{A}$ and shape $G$.
As we see, PBIDR captures the illumination environment in each input image, and can successfully transfer both the diffuse and specular reflection between these cases.
In Fig.~\ref{suppfig: relighting_novel},
we introduce two significantly different light sources,
with up-left and down-right directions
and use these to render them.
PBIDR can successfully decompose these synthetic lights and transfer them between subjects.

\section{Expression Manipulation}
\label{sec4}
\begin{figure}[ht]
    \centering
    \includegraphics[width=\linewidth]{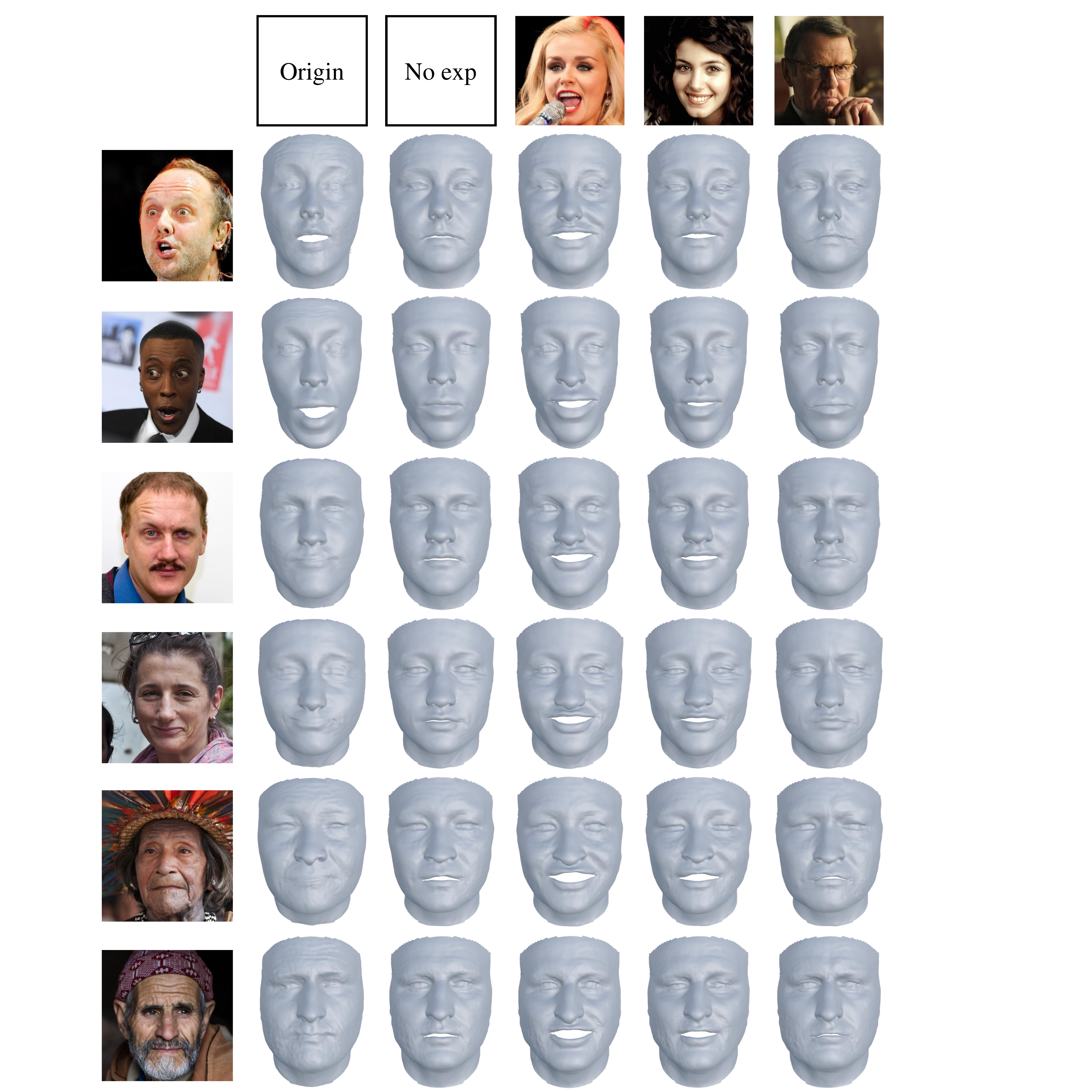}
    \caption{Expression manipulation of our detailed mesh reconstructions.
    On main advantage is that the details can be transferred on a 3DMM reconstruction of the same subject, to create a more detailed 3DMM.
    Here, we manipulate such a model's blenshapes to introduce expressions from the images of the first row.}
    \label{suppfig: expression manipulation}
\end{figure}

We represent details as displacements based on mesh vertices, which allows the inferred details to be transferred to a mesh,
regardless of the expression changes.
In Fig. \ref{suppfig: expression manipulation}, we use three different expression blendshapes to manipulate our detailed meshes. We find that even with significant expression changes, our mesh retains details while being manipulated.

\section{Additional Results}
\label{sec5}

\begin{figure}
    \centering
    \includegraphics[width=0.9\linewidth]{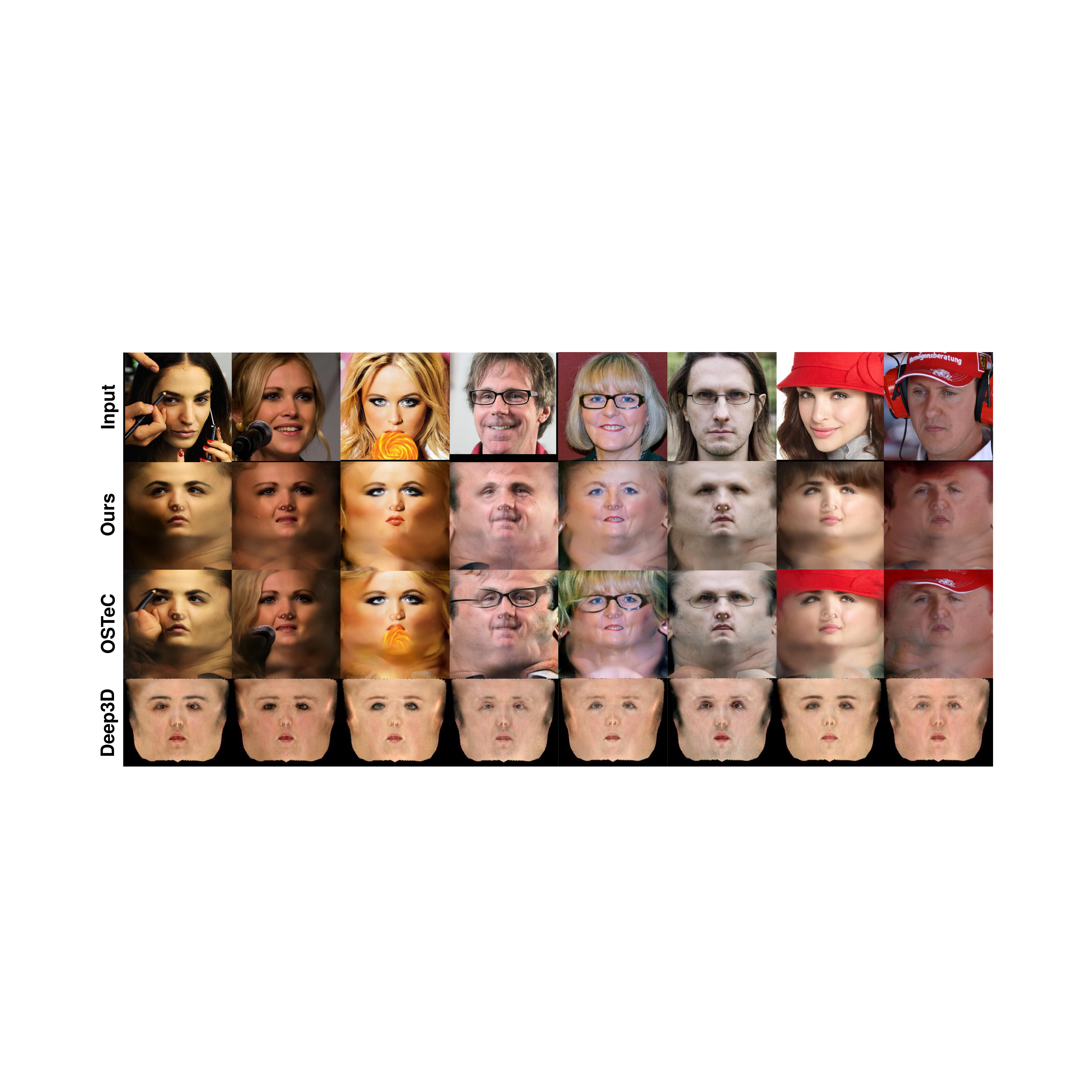}
    \vspace{-9mm}
    \caption{Additional texture-completion and de-occlusion results compared to OSTeC \cite{gecer2021ostec} and Deep3D \cite{deng2019accurate}, including an example failure case in subjects wearing hats, where the occlusion is replaced with hair.}
    \vspace{-9mm}
    \label{suppfig: texture completion}
\end{figure}

\begin{figure}
    \centering
    \includegraphics[width=0.9\linewidth]{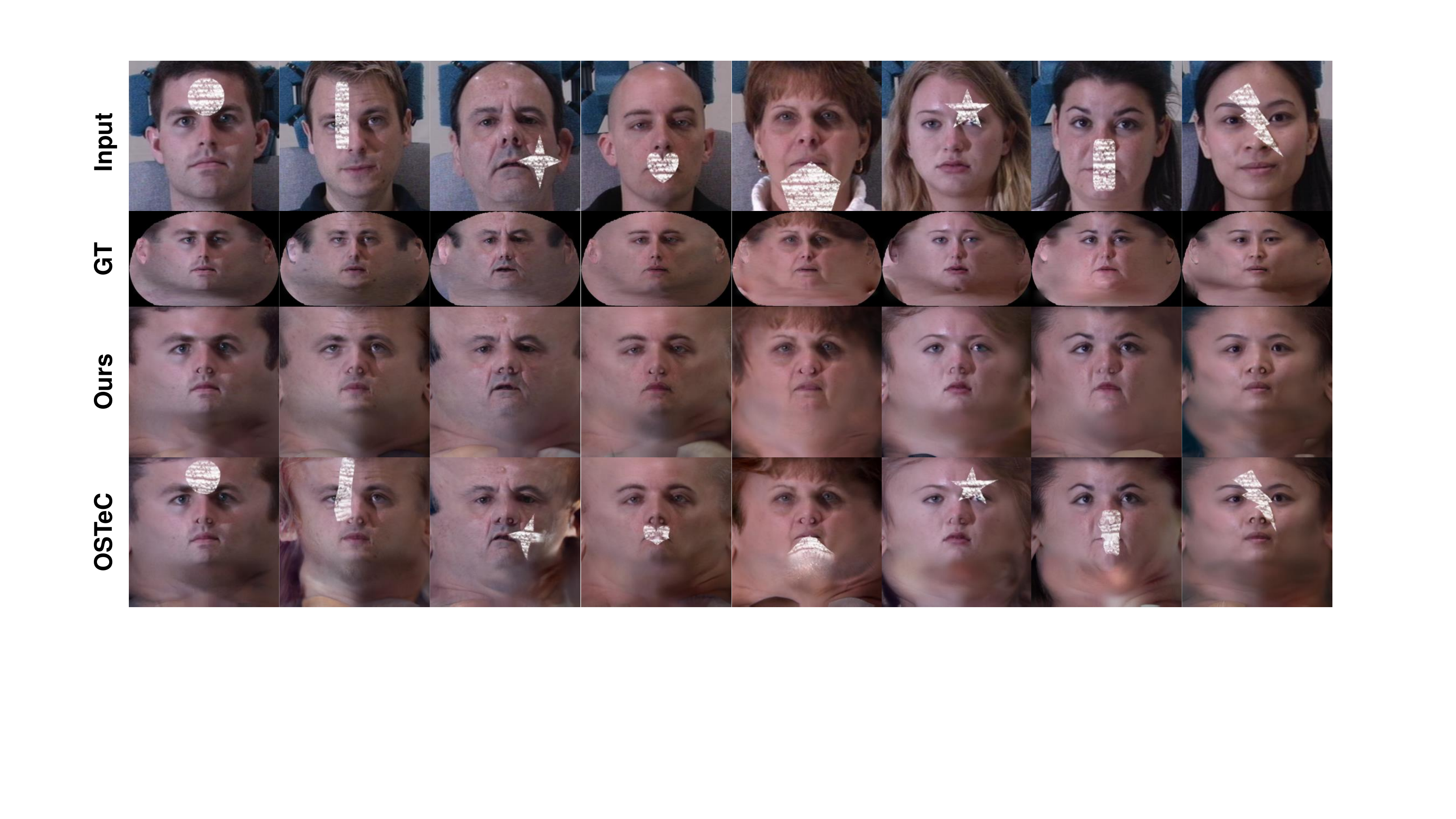}
    \vspace{-6mm}
    \caption{Comparison our method with texture completion methods in Multi-PIE~\cite{gross2010multipie}. The occlusions are inserted manually. Results exhibit that our approach can remove occlusions and complete photo-realistic textures.}
    \vspace{-6mm}
    \label{suppfig: uvdb}
\end{figure}

We show additional texture completion results in Fig.~\ref{suppfig: texture completion}. Consistent with the main manuscript, our method achieves high quality and photo-realistic textures. We also depict some failure cases, where the occlusions are not entirely removed, mainly because of improper segmentation. For example, in the last two columns where subjects wear hats, our method removes some of the occlusion but still replaces it with hair. In Fig.~\ref{suppfig: uvdb}, we further qualitatively depict examples of occlusion in the manuscript Tab.~\textcolor{red}{4}. The occlusions will cause artifacts in other texture completion algorithms, while ours can remove the occlusion stably.
\end{document}